\documentclass{article} 
\usepackage{iclr2027_conference,times}

\usepackage{amsmath,amsfonts,bm}

\def\eqref#1{equation~\ref{#1}}

\def\1{\bm{1}}

\DeclareMathAlphabet{\mathsfit}{\encodingdefault}{\sfdefault}{m}{sl}
\SetMathAlphabet{\mathsfit}{bold}{\encodingdefault}{\sfdefault}{bx}{n}

\usepackage{hyperref}
\usepackage{url}
\usepackage{booktabs}
\usepackage{array}
\usepackage{multirow}
\usepackage{amssymb}
\usepackage{amsmath}
\usepackage{graphicx}

\usepackage{longtable}
\usepackage{cuted}
\usepackage{enumitem}
\usepackage{tabularx}
\usepackage[misc]{ifsym}

\usepackage{caption}
\title{TacMind: Benchmarking and Modeling Tactile Interaction for Robotic Manipulation}

\author{
    Jiaxin Shi$^{1,4}$ \;
    Zibin Wu$^{2,4}$ \;
    Yichi Zhang$^{3,4}$ \; 
    \textbf{Jianghua Duan$^{4}$} \;
    \textbf{Siyu Zhu$^{5}$} \;
    \textbf{Weihao Yuan$^{6,4,}$\textsuperscript{\Letter}} \\[4pt]
    \textsuperscript{1}Shanghai Jiao Tong University 
    \textsuperscript{2}The Hong Kong University of Science and Technology (Guangzhou) \\
    \textsuperscript{3}Harbin Institute of Technology \;
    \textsuperscript{4}Daimon Robotics \;
    \textsuperscript{5}Fudan University \;
    \textsuperscript{6}Nanjing University \\[2pt]
    \textsuperscript{\Letter} Corresponding Author
}

\iclrfinalcopy 
\begin{document}

\maketitle
\fancyhead{}                      

\begin{figure}[h]
\centering
\vspace{-8mm}
\includegraphics[width=\linewidth]{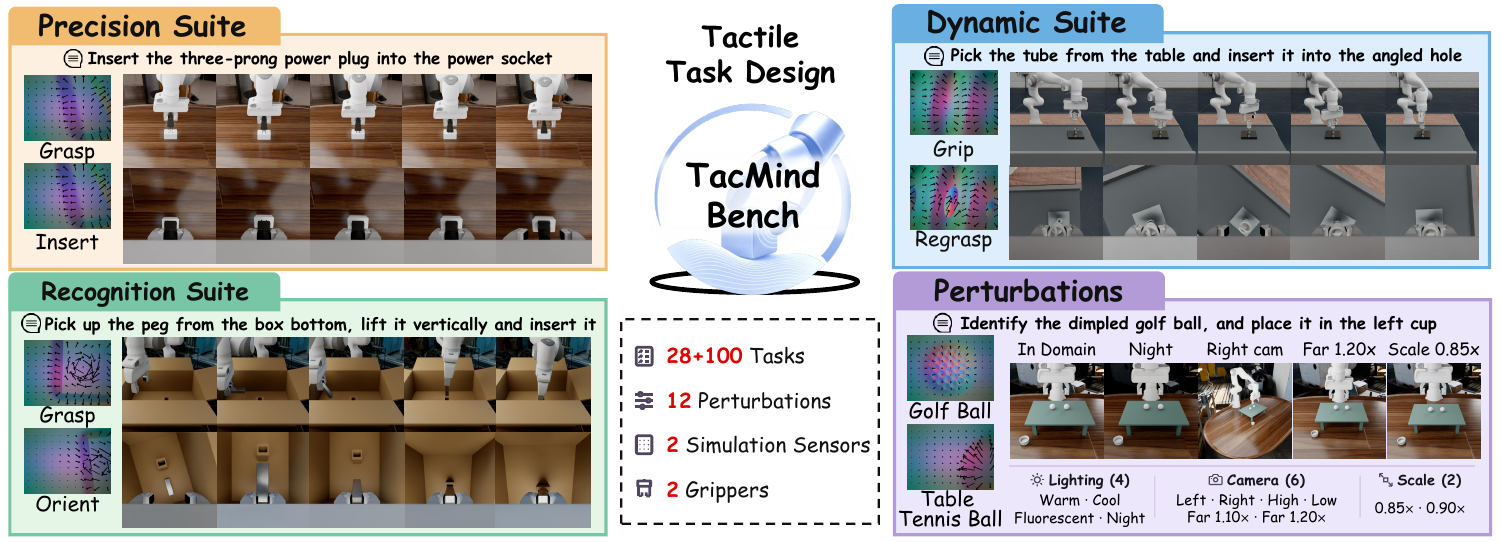}
\captionsetup{hypcap=false}
\captionof{figure}{
\textbf{Overview of \textsc{TacMind-Bench}.} \textsc{TacMind-Bench} comprises 28 Standard evaluation tasks and 100 Extra demonstration tasks across precise contact, recognition under ambiguity, and dynamic interaction, with 12 controlled perturbations and two tactile sensor embodiments.
}
\vspace{3mm}
\label{fig:teaser}
\end{figure}
\begin{abstract}
\vspace{-4mm}
Tactile sensing reveals contact geometry and dynamics, yet contact-rich manipulation requires interpreting recent interactions and anticipating how candidate actions will change contact.
Existing evaluations largely isolate individual skills, while policies conditioned only on observed contact lack explicit forecasts of their actions' tactile consequences.
We introduce \textsc{TacMind-Bench}, a simulation benchmark with 28 Standard evaluation tasks and 100 Extra demonstration tasks across 287 object assets and 11 contact behaviors.
Organized around precise contact, recognition under ambiguity, and dynamic interaction, it provides demonstrations and 12 controlled perturbation settings across lighting, camera viewpoint, and object scale.
To connect contact understanding with action generation, we develop the Tac-LeWM latent world model and the TacMind-0 vision-tactile-language-action (VTLA) policy.
Tac-LeWM learns a streaming tactile interaction state through action-conditioned prediction of future tactile embeddings, with canonical tactile rotary encoding and causal caching to represent spatial structure and recent history.
TacMind-0 uses this state to generate an initial action proposal, then applies predictive tactile reflection to refine the proposal through flow reversal conditioned on its predicted tactile consequences.
On the 28 Standard simulation tasks, TacMind-0 outperforms existing multi-task baselines with success rates of 34.1\% in-domain and 20.2\% under perturbations.
Experiments on six real-world robot subtasks further validate the approach on physical contact-rich manipulation. Project Page: \href{https://tacmind-bench.github.io/}{https://tacmind-bench.github.io/}.
\end{abstract}

\section{Introduction}

Tactile sensing complements vision with direct physical measurements of local contact geometry, elastomer deformation, and slip~\cite{howe1993tactile,hogan2020tactile}.
Visuo-tactile policies use these cues for tight-tolerance insertion under visual occlusion, and for manipulation when object pose, weight, or surface texture is visually ambiguous~\cite{3dvitac,manifeel,univtac,omnivta,wu2025tacdiffusion}.
Contact evolves continuously as manipulation proceeds through grasping, transport, alignment, and insertion, so its interpretation depends on recent tactile history.
Similar instantaneous deformation patterns can result from different preceding motions.
However, conditioning action chunks only on observed contact leaves the policy without an explicit forecast of how its proposed motions will change contact or trigger slip.

Existing tactile benchmarks have standardized the evaluation of individual manipulation skills, but mostly within isolated settings and narrow test conditions~\cite{maniskillvitac,manifeel,univtac,tabero,softvtbench}.
Initializing objects in hand near the target isolates terminal alignment, yet omits the grasp offsets introduced during pickup that can compound into severe jamming during insertion.
Standalone probing evaluates tactile recognition separately from the downstream manipulation it must guide; stationary setups likewise omit reactive alignment under externally driven motion.
Without orthogonal visual and physical perturbations, evaluations also leave unclear whether tactile feedback compensates for degraded visual observations or altered contact clearances.
These challenges call for a unified benchmark of diverse interaction regimes and a predictive tactile representation that couples historical context with future contact anticipation.

We introduce \textsc{TacMind-Bench}, a simulation benchmark organized by the main functional role of touch as shown in Fig.~\ref{fig:teaser}.
Its three evaluation regimes cover \emph{precise contact} for high-precision insertion and mating, \emph{recognition under ambiguity} for identifying occluded or visually indistinguishable objects, and \emph{dynamic interaction} with externally driven source fixtures or targets.
The benchmark comprises 28 Standard evaluation tasks and 100 Extra demonstration tasks across 287 object assets and 11 contact behaviors.
It spans terminal alignment and full-horizon manipulation from object acquisition to placement, with static and dynamic task counterparts paired under shared geometries.
The Standard suite tests out-of-domain (OOD) robustness through 12 controlled perturbations of lighting, camera viewpoint, and object scale.
The Extra suite broadens the behavior repertoire for large-scale representation learning through tool-based cutting, surface wiping, rolling, and articulated mechanism manipulation.

Tac-LeWM learns a streaming tactile interaction state by predicting contact transitions across these regimes, with pretraining on over 3M frames of real-world and simulated multi-sensor demonstrations (Fig.~\ref{fig:taclewm}(c)).
Building on LeWM~\cite{leworldmodel}, we jointly train a latent encoder and predictor to forecast future tactile embeddings from recent history and candidate actions.
An action-reliance margin compares predictions under true and permuted action sequences to encourage sensitivity to the commanded motion.
Our Canonical Tactile Rotary (CTR) positional encoding maps mirrored dual-finger layouts into a shared coordinate frame, where multidirectional rotary encoding~\cite{liu2026spiral} represents non-axis-aligned shear and torsion fields.
A streaming causal encoder preserves temporal context through cached key-value (KV) features, so each incoming frame updates the state without re-encoding its predecessors.

TacMind-0 uses the current and predicted tactile states at successive stages of action generation within a multi-task vision-tactile-language-action (VTLA) policy, as shown in Fig.~\ref{fig:taclewm}(a).
Built on a pretrained vision-language-action (VLA) backbone and a flow-matching action expert, it first generates an action chunk from the streaming tactile state alongside vision and language inputs.
We then introduce \emph{predictive tactile reflection}: Tac-LeWM forecasts the future tactile latent state induced by this initial proposal, as shown in Fig.~\ref{fig:taclewm}(b).
The policy first inverts the initial flow to recover its latent noise anchor.
The predicted tactile state modulates the action expert through adaptive root-mean-square normalization (AdaRMSNorm), and flow-reversal steering regenerates the chunk from this anchor.
This extends the tactile-token conditioning used by N0-VTLA~\cite{team2026n_0} with prediction-conditioned action refinement.

Experiments on the 28 Standard tasks of TacMind-Bench demonstrate that TacMind-0 outperforms other baselines to achieve 34.1\% in-domain (InD) and 20.2\% out-of-domain (OOD) success. Ablations confirm the benefit of predictive tactile modeling, as streaming conditioning lifts InD success from 27.7\% to 31.3\%, while predictive reflection raises OOD success from 18.7\% to 20.2\% and further improves OOD precision from 6.7\% to 8.1\%. Moreover, linear probes show that frozen Tac-LeWM states encode proprioception, gripper state, actions, and pad deformation energy with Pearson correlations from 0.80 to 0.91, while multi-sensor pretraining, causal attention, and CTR encoding enable 34.42 dB reconstruction and 28.98 dB future prediction PSNR.

Our main contributions are as follows.
\begin{itemize}[leftmargin=*, topsep=0pt, parsep=2pt, itemsep=2pt, partopsep=0pt]
    \item We introduce \textsc{TacMind-Bench}, a simulation benchmark of precise contact, recognition under ambiguity, and dynamic interaction, with 28 Standard tasks, 12 controlled perturbations, and 100 Extra tasks spanning 11 contact behaviors.

    \item We develop Tac-LeWM to learn streaming tactile interaction states through action-conditioned latent prediction, with CTR positional encoding for dual-finger contact geometry and causal KV caching for incremental state updates.

    \item We propose TacMind-0, a multi-task VTLA policy that conditions initial action generation on the current tactile state and refines the proposal through predictive tactile reflection and flow-reversal steering.
\end{itemize}

\section{Related Work}

\noindent\textbf{Tactile Manipulation Benchmarks.}
Tactile simulators support scalable development and evaluation of visuo-tactile manipulation systems~\cite{tacto,taxim,tacsl,tacex}.
Recent suites standardize contact-rich control and deformable manipulation~\cite{softvtbench}, with complementary emphases on different physical interactions.
ManiSkill-ViTac~\cite{maniskillvitac} isolates terminal alignment with objects initialized in hand; ManiFeel~\cite{manifeel} and UniVTAC~\cite{univtac} cover static tasks with partial end-to-end acquisition; and Tabero~\cite{tabero} focuses on force regulation.
\textsc{TacMind-Bench} brings full-horizon acquisition, state disambiguation, and externally driven motion into a common evaluation organized by the functional role of touch.
Its lighting, viewpoint, and object-scale perturbations provide controlled tests of robustness across these interaction regimes.

\noindent\textbf{Visuo-Tactile Policy Learning and Predictive Modeling.}
Research on transferable tactile representations~\cite{tdex,unit,sparsh,t3,tacx,feng2026anytouch} has complemented policy learning for dexterous and contact-rich manipulation~\cite{rotateit,robotsynesthesia,mimictouch,3dvitac,wu2025tacdiffusion}.
Building on diffusion and flow-matching policies~\cite{chi2025diffusion,pi05,opendm2026}, recent methods integrate tactile feedback into visuomotor and VLA models~\cite{rdp,vlatouch,team2026n_0,unitacvla,omnivta}.
These developments motivate representations that account explicitly for cross-fingertip mirror symmetry and the directional structure of contact deformation.
Predictive tactile models further anticipate contact through action-conditioned representations, future visuo-tactile rollouts, and subgoals~\cite{vtwm,touchworld,vitacworld}, or use contact-state and slip forecasts for action ranking and reactive correction~\cite{tacforesight,hitacwam,unitacvla,omnivta}.
Tac-LeWM couples canonical spatial encoding with causal KV caching to maintain a streaming tactile interaction state.
Its action-conditioned predictor forecasts future tactile latents without reconstructing future pixels.
TacMind-0 uses these forecasts to modulate continuous action generation through flow reversal, rather than using them only to select among completed proposals.

\section{TacMind-Bench Design}

\begin{figure*}[t]
    \centering
    \includegraphics[width=\textwidth]{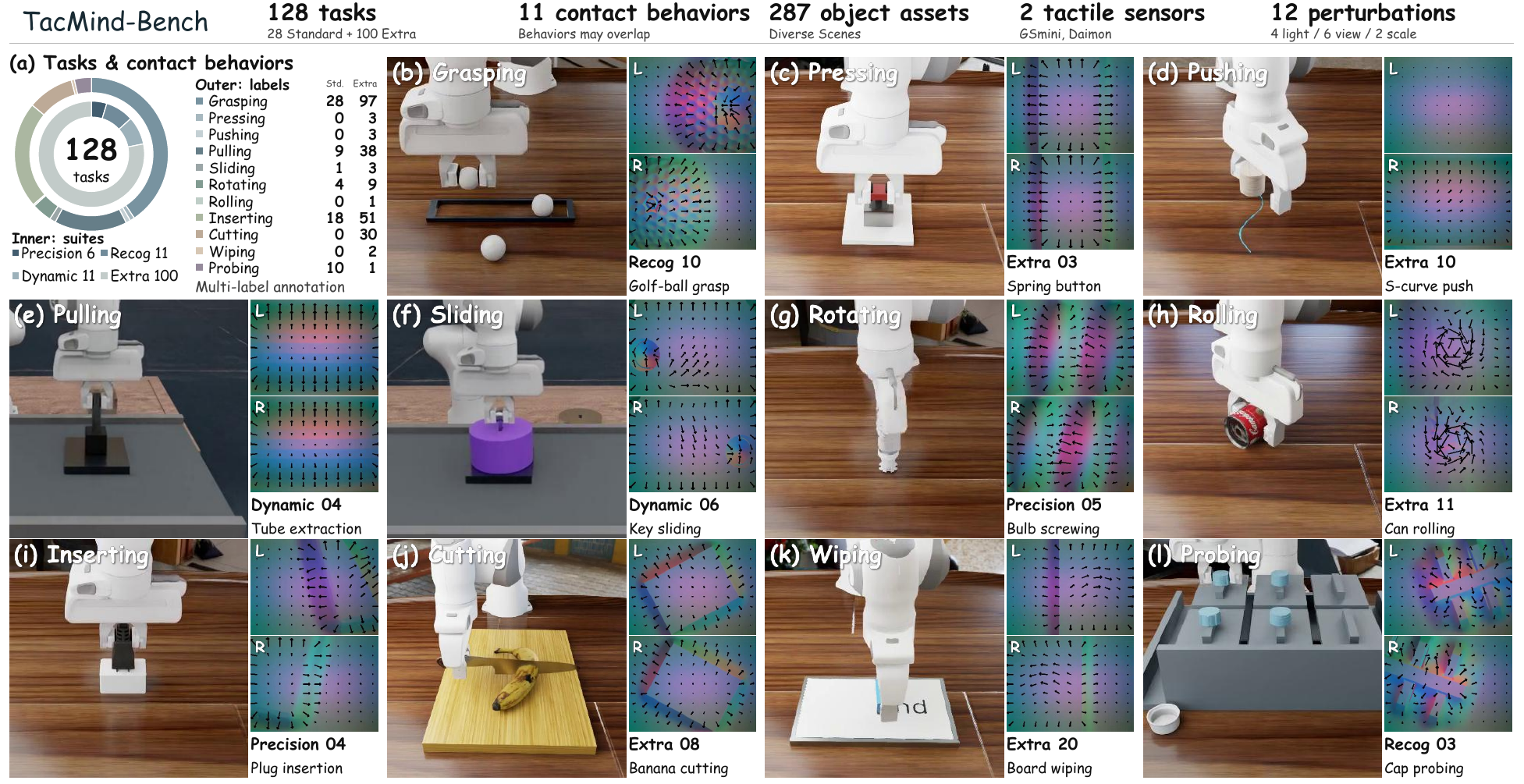}
    \caption{
    \textbf{\textsc{TacMind-Bench} Tasks.}
    \textbf{(a)} Task composition across the 28 Standard evaluation tasks and 100 Extra demonstration tasks (inner ring) and multi-label annotations across 11 contact behaviors (outer ring and table).
    \textbf{(b-l)} Representative tasks across the 11 contact behaviors paired with tactile visualizations from the left L and right R fingers.
    }
    \label{fig:tacmind_bench_overview}
\end{figure*}

\textsc{TacMind-Bench} organizes tasks by the role tactile information plays in successful manipulation, rather than by object category or manipulation primitive.
The \textbf{Standard} suite contains 28 curated tasks for multi-task and robustness evaluation, while the \textbf{Extra} suite provides 100 demonstration tasks for representation learning and skill extension.
Across 287 object assets, we provide environments and expert demonstrations for two vision-based tactile sensors with complementary sensing properties.
GelSight Mini captures fine local geometry and surface texture; Daimon emphasizes dynamic measurements of normal pressure, tangential shear, and slip.
The Standard tasks fall into three regimes: \textbf{precise contact}, \textbf{recognition under ambiguity}, and \textbf{dynamic interaction}, as shown in Fig.~\ref{fig:tacmind_bench_overview} (a).
We annotate all tasks with 11 contact-behavior labels, allowing multiple labels because a task can pass through several contact phases.
Fig.~\ref{fig:tacmind_bench_overview} (b) to (l) illustrates these behaviors through paired scene and dual-finger tactile observations; Table~\ref{tab:tactile_benchmark_comparison} compares existing suites along four design axes.

\begin{table}[t]
\centering
\caption{
\textbf{Comparison with tactile manipulation benchmarks.}
$\checkmark$, $\triangle$, and $\times$ denote explicit, partial, and omitted coverage across full-horizon object acquisition (\textbf{E2E}), state disambiguation (\textbf{SD.}), dynamic motion (\textbf{Dyn.}), and orthogonal perturbations (\textbf{Perturb.}).
}
\label{tab:tactile_benchmark_comparison}
\small
\setlength{\tabcolsep}{14pt}
\begin{tabular}{l| c | c c c c}
\toprule
\textbf{Benchmark}
& \textbf{Tasks}
& \textbf{E2E}
& \textbf{SD.}
& \textbf{Dyn.}
& \textbf{Perturb.} \\
\midrule
ManiSkill-ViTac~\cite{maniskillvitac}
& 3
& $\times$ & $\times$ & $\times$ & $\times$ \\
ManiFeel~\cite{manifeel}
& 9
& $\triangle$ & $\triangle$ & $\times$ & $\times$ \\
UniVTAC~\cite{univtac}
& 8
& $\triangle$ & $\triangle$ & $\times$ & $\times$ \\
Tabero~\cite{tabero}
& 40
& $\checkmark$ & $\times$ & $\times$ & $\times$ \\
\textbf{\textsc{TacMind-Bench}}
& \textbf{28+100}
& $\checkmark$
& $\checkmark$
& $\checkmark$
& $\checkmark$ \\
\bottomrule
\end{tabular}
\vspace{-4mm}
\end{table}

\noindent\textbf{Precise Contact.}
The 6 Standard tasks evaluate precise insertion and mating in static scenes.
Vision provides coarse alignment, while tactile observations reveal local jamming and multidirectional shear at the contact interface.
Two tasks isolate terminal alignment by starting with the object in hand: screwing a bulb and inserting a power plug.
The remaining 4 tasks require pickup, transport, contact establishment, and insertion of tubes, keys, connectors, or gears.
This distinction exposes how small grasp offsets acquired during pickup can propagate into jamming during insertion.

\noindent\textbf{Recognition under Ambiguity.}
These 11 Standard tasks require policies to infer object pose, size, surface texture, rib orientation, or boundary contour when vision leaves these properties ambiguous.
7 tasks conceal objects in covered trays or boxes, with examples including rectangular-peg extraction and cap or token probing.
Four tasks place visually indistinguishable objects in open view: cylinders with different thread densities, caps with horizontal or vertical ribs, and dimpled golf balls versus smooth table-tennis balls.
Physical exploration must inform the subsequent grasp, reorientation, or placement, so recognition is evaluated as part of manipulation under occlusion, limited viewpoints, and clutter.

\noindent\textbf{Dynamic Interaction.}
The 11 Standard dynamic tasks introduce externally driven conveyor motion into precision manipulation through three settings.
In \emph{moving-target alignment}, 3 tasks begin with the object in hand while the target socket translates, separating online alignment from object acquisition.
Four \emph{static-source to moving-target} tasks retain the complete pick-and-insert sequence while the target fixture moves continuously.
The other 4 tasks reverse this arrangement: in \emph{moving-source to static-target} manipulation, the robot acquires an object from a moving fixture and inserts it into a stationary target.
Examples include dynamic tube extraction and key sliding.
Static and dynamic counterparts share the same object, target, and manipulation primitive, supporting controlled comparisons of stationary and time-varying contact.
Continuous translation narrows the time available for correction, so delayed responses to slip or alignment error can cause failure.

\noindent\textbf{Controlled Perturbations and Extra Suite.}
In addition to randomizing initial object positions, we evaluate 12 controlled perturbations across the three regimes: 4 illumination conditions, 6 camera-viewpoint shifts, and 2 object-scale variations.
Lighting and viewpoint changes modify visual observations while preserving tactile sensing conditions, testing the complementary role of touch when visual input changes.
Object-scale changes instead alter contact clearances and challenge action sequences that do not adapt to the resulting jamming or slip.
The 100 Extra tasks broaden demonstration coverage across all 11 contact behaviors.
They comprise 27 articulated-mechanism, non-prehensile, and surface-interaction tasks, including pressing, pushing, rolling, and wiping; 30 tool-based cutting tasks with varied food geometries; and 43 compositional assembly tasks with diverse materials and multi-target layouts.

\noindent\textbf{Task Difficulty and Agentic Data Generation.}
Strict completion criteria, such as full plug insertion and complete gear meshing, combine with multi-stage horizons, moving fixtures, and geometric perturbations to expose policy bottlenecks.
These requirements also motivate precise demonstrations with minimal teleoperation jitter.
We therefore develop a closed-loop agentic pipeline on Isaac Sim 5.1~\cite{NVIDIA_Isaac_Sim} and CuRobo~\cite{curobo} for both GelSight Mini and Daimon gripper configurations.
The Appendix details the collection and verification procedures.

\begin{figure*}[t]
\centering
\includegraphics[width=\linewidth]{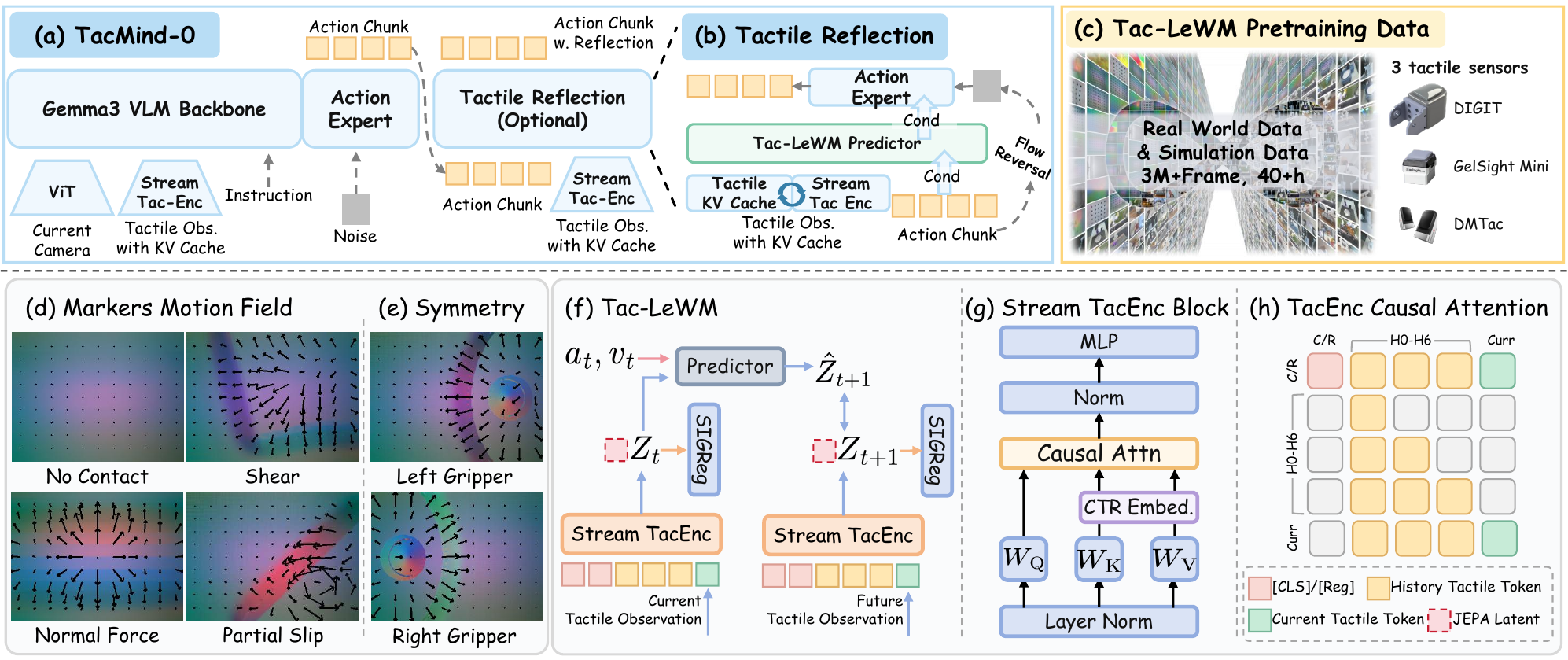}
\caption{
\textbf{Overview of TacMind-0 and Tac-LeWM.}
\textbf{(a)} TacMind-0 maintains a streaming tactile interaction state via causal key-value caching for initial action chunk generation.
\textbf{(b)} Predictive tactile reflection uses Tac-LeWM to forecast action-conditioned future tactile states and refines candidate actions via flow-reversal steering.
\textbf{(c)} Tac-LeWM is pretrained on over 3M frames and 40 hours of real-world and simulated demonstrations across three tactile sensors.
\textbf{(d, e)} Marker-motion directions vary across contact conditions while the left and right fingertip sensors have mirrored spatial layouts, motivating directional and symmetry-aware spatial encoding.
\textbf{(f)} Tac-LeWM combines a streaming tactile encoder with an action-conditioned latent predictor to forecast future tactile embeddings.
\textbf{(g)} The streaming encoder reuses historical features through cached keys and values for efficient online updates.
\textbf{(h)} Causal attention integrates preceding tactile context while excluding future frames.
}
\label{fig:taclewm}
\vspace{-6mm}
\end{figure*}

\section{TacMind-0}

\noindent\textbf{Overview.}
TacMind-0 is a multi-task VTLA policy that maintains a predictive tactile interaction state to address history-dependent state disambiguation and contact-rich alignment in \textsc{TacMind-Bench}.
At time $t$, the policy takes visual observations $I_t$, a language instruction $L$, and recent tactile history $\mathcal{H}_t^\tau=(\tau_{t-H+1},\ldots,\tau_t)$ over context length $H$, and predicts an action chunk $A_t=[a_t,\ldots,a_{t+K-1}]$ over horizon $K$.
Tac-LeWM encodes $\mathcal{H}_t^\tau$ into a streaming tactile interaction state $z_t^\tau$ learned through action-conditioned prediction of future tactile latents.
As shown in Fig.~\ref{fig:taclewm}, TacMind-0 combines $z_t^\tau$ with vision and language to generate an initial action proposal $A_t^{(0)}$, predicts its future contact consequences, and refines the proposal through predictive tactile reflection:
\begin{equation}
\{I_t,\mathcal{H}_t^\tau,L\}
\xrightarrow{E_\psi,\,\pi_\theta}
A_t^{(0)}
\xrightarrow{P_\phi}
\hat{Z}_{t,\mathrm{ref}}^{+}
\xrightarrow{F_\theta^{-1},\,F_\theta}
A_t^{(*)},
\end{equation}
where $E_\psi$ and $P_\phi$ denote the Tac-LeWM streaming encoder and latent predictor, $\hat{Z}_{t,\mathrm{ref}}^{+}$ denotes the predicted future tactile latent chunk, and $F_\theta^{-1}$ followed by $F_\theta$ represents flow-reversal steering within the policy $\pi_\theta$.

\subsection{Learning a Predictive Tactile State}
Tac-LeWM adapts the joint-embedding predictive formulation of LeWM~\cite{leworldmodel} to vision-based tactile sensing, forecasting future latent embeddings rather than reconstructing pixels.

\noindent\textbf{Tactile Input Representation.}
For each finger $f\in\{L,R\}$, we overlay the marker-displacement field $\tau_t^{f,\mathrm{mark}}$ onto the tactile RGB image $\tau_t^{f,\mathrm{rgb}}$ to obtain $\tilde{\tau}_t^f=\mathcal{O}(\tau_t^{f,\mathrm{rgb}},\tau_t^{f,\mathrm{mark}})$, where $\mathcal{O}$ denotes the overlay operation.
The overlay combines local surface appearance and contact geometry with deformation cues from shear, torsion, and slip.
We then arrange the left and right fingertip observations into $\tau_t=\mathcal{M}(\tilde{\tau}_t^L,\tilde{\tau}_t^R)$ and patchify the result into tactile tokens.

\noindent\textbf{Canonical Tactile Rotary Positional Embedding.}
Standard image positional encodings associate tokens with global raster locations, while axial RoPE encodes horizontal and vertical offsets separately~\cite{heo2024rotary, liu2026spiral}.
Tactile sensing requires spatial encoding that accounts for both deformation direction and sensor mounting.
First, contact-induced motion need not follow the image axes: normal loading produces radial outward displacement, whereas tangential shear and partial slip induce oblique, nonuniform motion fields, as shown in Fig.~\ref{fig:taclewm} (d).
Second, the left and right sensors are mounted in mirrored orientations, whereas global raster locations depend on their placement in the input mosaic, as shown in Fig.~\ref{fig:taclewm} (e).
CTR therefore encodes position in canonical local fingertip coordinates, with $p_{f,t,y,x}=(t,y,x)$ denoting a token from finger $f$ at time $t$ and local patch coordinate $(y,x)$; a learnable finger embedding retains sensor identity.
We map the right fingertip into the shared frame via $x^{\mathrm{can}}=W_{\mathrm{cell}}-1-x^{\mathrm{raw}}$, where $W_{\mathrm{cell}}$ is the local patch-grid width, so that coordinate directions align across fingers without altering the input image.
Within this frame, we project local coordinates along eight orientations as $u_k(x,y)=x\cos\phi_k+y\sin\phi_k$, where $\phi_k=k\pi/8$ for $k\in\{0,\ldots,7\}$, and apply rotary transformations to the query and key features in tactile causal attention.
For tokens separated by $(\Delta x,\Delta y)$, the spatial rotary phase differences depend on $\Delta u_k=\Delta x\cos\phi_k+\Delta y\sin\phi_k$; these projections encode relative position along multiple directions across oblique deformation fields.

\noindent\textbf{Streaming Causal Encoding.}
To retain contact history as new observations arrive, we encode $\mathcal{H}_t^\tau$ with causal attention and reuse historical features through a KV cache as shown in Fig.~\ref{fig:taclewm} (g,h).
Let $Z_t^\tau=E_\psi(\tau_t;\mathrm{KV}_{<t})$ denote the tactile token features at time $t$, where $\mathrm{KV}_{<t}$ stores cached keys and values from preceding frames within the context horizon $H$.
Current tactile tokens attend to both current and cached historical tokens, with future observations excluded; the global token $z_t^\tau=\mathrm{CLS}(Z_t^\tau)$ summarizes this context as the streaming tactile interaction state.
At inference, reusing $\mathrm{KV}_{<t}$ updates $z_t^\tau$ incrementally from the newly arrived observation $\tau_t$ alone.
This avoids re-encoding preceding frames and bounds the per-step computation over a continuing interaction for a fixed context horizon.

\noindent\textbf{Action-Conditioned Latent Prediction.}
We train the streaming state to predict how contact evolves under an action sequence, as shown in Fig.~\ref{fig:taclewm} (f).
At time $t$, a future tactile chunk $\mathcal{T}_t^{+}=(\tau_{t+\Delta},\ldots,\tau_{t+\Delta+C-1})$ with prediction offset $\Delta$ and horizon $C$ defines latent targets $Z_t^{+}=(z_{t+\Delta}^\tau,\ldots,z_{t+\Delta+C-1}^\tau)$, which are produced directly by $E_\psi$.
Conditioned on the intervening actions $A_t^{+}=(a_t,\ldots,a_{t+\Delta+C-2})$, the predictor estimates
$
\hat{Z}_t^{+}
=
P_\phi\left(z_t^\tau,A_t^{+},v_t\right),
$
where $v_t$ is an auxiliary scene feature extracted by a frozen DINOv3~\cite{dinov3} encoder.
During training, $v_t$ helps disambiguate contact onset as the gripper approaches an object from free space, where the tactile signals are initially blank.
We randomly drop $v_t$ during Tac-LeWM training to support prediction without this auxiliary input, and omit it entirely during policy post-training and inference.
Online reflection thus conditions $P_\phi$ solely on the tactile state $z_t^\tau$ and candidate actions.

\noindent\textbf{Predictive Learning Objective.}
The latent prediction loss over horizon $C$ is defined as
\begin{equation}
\mathcal{L}_{\mathrm{pred}}
=
\frac{1}{C}
\sum_{j=0}^{C-1}
\left\|
\hat{z}_{t+\Delta+j}^\tau
-
z_{t+\Delta+j}^\tau
\right\|_2^2.
\end{equation}
To prevent representation collapse during joint optimization of $E_\psi$ and $P_\phi$, we apply  SIGReg~\cite{leworldmodel} to the tactile embeddings $Z^\tau$ across time steps, yielding the overall objective
\begin{equation}
\mathcal{L}_{\mathrm{Tac\text{-}LeWM}}
=
\mathcal{L}_{\mathrm{pred}}
+
\lambda\,\mathrm{SIGReg}(Z^\tau),
\end{equation}
where $\lambda$ controls the regularization weight.

\begin{table*}[t]
    \centering
    \caption{\textbf{Performance on \textsc{TacMind-Bench} Standard (GelSight Mini).}
    OOD is categorized into lighting, camera, and size perturbations.}
    \vspace{-3mm}
    \label{tab:main_results}
    \small
    \setlength{\tabcolsep}{4pt}
    \begin{tabular}{l*{14}{c}}
        \toprule
        \multirow{2}{*}{Model}
        & \multicolumn{4}{c}{Precision}
        & \multicolumn{4}{c}{Recognition}
        & \multicolumn{4}{c}{Dynamic}
        & \multicolumn{2}{c}{Avg} \\
        \cmidrule(lr){2-5}
        \cmidrule(lr){6-9}
        \cmidrule(lr){10-13}
        \cmidrule(lr){14-15}
        & InD & Light & Cam & Size
        & InD & Light & Cam & Size
        & InD & Light & Cam & Size
        & InD & OOD \\
        \midrule

        ACT
        & 16.7 & 8.3 & 2.6 & 6.7
        & 19.6 & 13.1 & 7.0 & 11.6
        & \textbf{20.0} & 2.5 & 0.8 & 2.0
        & 19.1 & 5.6 \\

        OpenPi0.5
        & 12.5 & 4.8 & 2.8 & 9.6
        & 47.3 & 43.6 & \textbf{46.4} & 15.9
        & 3.2 & 1.8 & 2.5 & 2.0
        & 22.5 & 17.7 \\

        OpenDM
        & 10.8 & 7.9 & 3.6 & 7.9
        & 48.2 & 52.6 & 27.3 & \textbf{29.6}
        & 16.4 & 13.2 & 3.8 & \textbf{3.6}
        & 27.7 & 18.1 \\

        N0-VTLA
        & 6.7 & 3.3 & 1.3 & 7.9
        & 22.7 & 17.7 & 15.4 & 20.2
        & 3.6 & 4.2 & 1.2 & 0.7
        & 11.8 & 8.2 \\

        \textbf{TacMind-0}
        & \textbf{20.8} & \textbf{12.1}
        & \textbf{4.9} & \textbf{10.0}
        & \textbf{56.4} & \textbf{53.3}
        & 30.7 & 29.5
        & 19.1 & \textbf{18.2}
        & \textbf{4.8} & 3.2
        & \textbf{34.1} & \textbf{20.2} \\

        \bottomrule
    \end{tabular}%
    \vspace{-3mm}
\end{table*}

\subsection{Tactile-Conditioned Action Generation}

\noindent\textbf{VTLA Architecture.}
TacMind-0 incorporates the streaming tactile state into a pretrained VLA backbone with Gemma~3 for multimodal encoding and a $\pi_{0.5}$-style~\cite{pi05} flow-matching action expert, as shown in Fig.~\ref{fig:taclewm} (a).
Given the visual observation $I_t$, language instruction $L$, and streaming tactile state $z_t^\tau$, we project $z_t^\tau$ to the action-expert hidden dimension as tactile conditioning tokens and form the multimodal conditioning $c_t=\mathcal{F}(I_t,L,z_t^\tau)$.
The resulting policy generates an initial action chunk $A_t^{(0)}=\pi_\theta(c_t)$, where each action $a_t\in\mathbb{R}^{8}$ comprises a $7$-DoF arm joint-position command and a continuous gripper command.
To train the action expert, we interpolate between Gaussian noise and a demonstrated action chunk: given $A_t^{\mathrm{gt}}$ and $\epsilon\sim\mathcal{N}(0,I)$, the interpolant is $A_s=(1-s)\epsilon+sA_t^{\mathrm{gt}}$ at flow time $s\in[0,1]$, with target velocity $u_s=A_t^{\mathrm{gt}}-\epsilon$.
The flow-matching objective is
\begin{equation}
\mathcal{L}_{\mathrm{FM}}
=
\mathbb{E}
\left[
\left\|
v_\theta(A_s,s\mid c_t)-u_s
\right\|_2^2
\right].
\end{equation}
At inference, integrating $v_\theta$ from $\epsilon$ produces the initial action proposal $A_t^{(0)}$.

\noindent\textbf{Predictive Tactile Reflection.}
The state $z_t^\tau$ summarizes observed history up to time $t$, but the initial generation of $A_t^{(0)}$ does not explicitly incorporate the future contact state that executing $A_t^{(0)}$ would induce.
Predictive tactile reflection feeds these anticipated consequences back into action generation, as shown in Fig.~\ref{fig:taclewm} (b).
Given $A_t^{(0)}$, Tac-LeWM takes the action prefix matching horizon $C$ to predict the future tactile latent chunk $\hat{Z}_{t,\mathrm{ref}}^{+}=P_\phi(z_t^\tau,A_t^{(0)})$, which we pool temporally into $\hat{z}_{t,\mathrm{ref}}^{+}=\mathrm{Pool}(\hat{Z}_{t,\mathrm{ref}}^{+})$.
We refine $A_t^{(0)}$ through Flow Reversal Steering (FRS)~\cite{tang2026improvingroboticgeneralistpolicies}, which maps a reference action back into the latent noise space to leverage the learned action prior while preserving its underlying motion structure. Let $F_\theta(\cdot\mid c)$ denote the ODE flow mapping from latent noise to an action chunk. We numerically invert the base flow under $c_t$ to obtain a reference-aligned noise anchor $\tilde{A}_0=F_\theta^{-1}(A_t^{(0)}\mid c_t)$. Due to finite-step integration, this anchor need not coincide with the original sampled noise and may lie in a higher-density region of the training-time noise prior. We then inject $\hat{z}_{t,\mathrm{ref}}^{+}$ into the action-expert layers through adaptive root-mean-square normalization (AdaRMSNorm) to form the reflective conditioning $c_t^{\mathrm{ref}}=\mathcal{G}(c_t,\hat{z}_{t,\mathrm{ref}}^{+})$. Integrating forward from the reference-aligned anchor $\tilde{A}_0$ yields the refined action chunk $A_t^{(*)}=F_\theta(\tilde{A}_0\mid c_t^{\mathrm{ref}})$, allowing predicted tactile consequences to guide action refinement while retaining the initial proposal's overall motion structure.

\subsection{Training and Online Execution}

\noindent\textbf{Tac-LeWM Pretraining and Fine-tuning.}
We first pretrain $E_\psi$ and $P_\phi$ on the heterogeneous multi-sensor corpus in Fig.~\ref{fig:taclewm} (c), then fine-tune them on in-domain training demonstrations from the downstream tasks.
Both stages exclude all evaluation rollouts and OOD perturbation settings.
During pretraining, an action-reliance margin regularizer encourages predictions conditioned on the true action sequence to match the future tactile targets more closely than predictions conditioned on permuted actions from the same dataset, so that $P_\phi$ remains responsive to its action inputs.
Because the source datasets use different control parameterizations, lightweight dataset-specific MLP projections map their actions into a shared space before a common action encoder; the Appendix gives the complete objective and data protocol.

\noindent\textbf{VLA Post-Training.}
Post-training on the downstream demonstrations first learns tactile-conditioned generation, then predictive tactile reflection.
In the first stage, we freeze the fine-tuned tactile encoder $E_\psi$ to provide $z_t^\tau$ and optimize the multimodal projection and base action expert with $\mathcal{L}_{\mathrm{FM}}$ conditioned on $c_t$.
In the second stage, we freeze $E_\psi$, the VLM backbone, and the base action-expert weights, and jointly optimize only the predictor $P_\phi$ and the AdaRMSNorm modulation layers.
This stage combines the refined flow-matching loss $\mathcal{L}_{\mathrm{FM}}$ conditioned on $c_t^{\mathrm{ref}}$ with the latent prediction loss $\mathcal{L}_{\mathrm{pred}}$.
Keeping the base action expert fixed preserves its single-pass generation and inversion mapping $F_\theta(\cdot\mid c_t)$, while the lightweight AdaRMSNorm layers learn to modulate the forward trajectory in response to $\hat{z}_{t,\mathrm{ref}}^{+}$.

\noindent\textbf{Online Execution.}
As each tactile observation arrives, $E_\psi$ updates $z_t^\tau$ through its KV cache in 10.33\,ms per step; TacMind-0 then generates $A_t^{(0)}$ and refines it into $A_t^{(*)}$ through latent prediction and flow reversal.
Compiling the action expert with TensorRT and CUDA Graphs reduces the latency of the three-pass reflective integration to 14.0\,ms for closed-loop reflective control, as detailed in the Appendix.
\section{Experiments}

\begin{figure}[t]
\centering
\includegraphics[width=\linewidth]{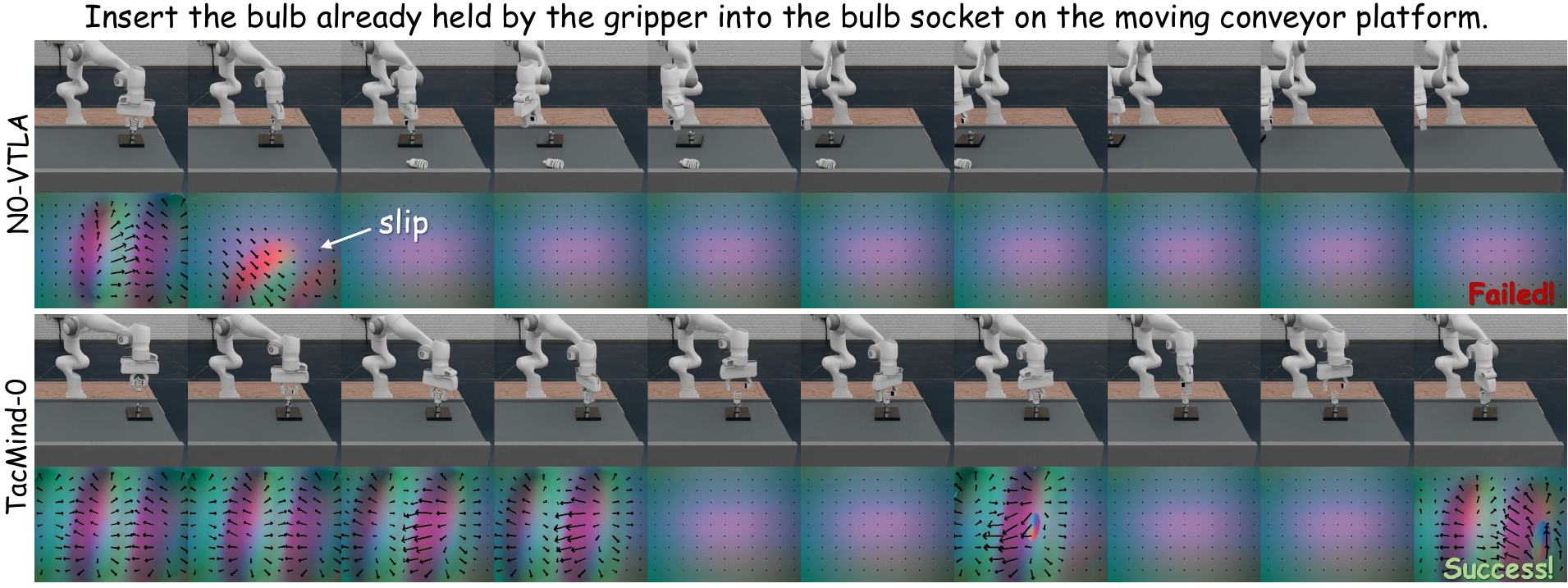}
\vspace{-6mm}
\caption{\textbf{TacMind-0 corrects contact during dynamic insertion.} A qualitative comparison with N0-VTLA on the Dynamic suite of \textsc{TacMind-Bench} shows bulb insertion into a moving socket.}
\label{fig:case1}
\vspace{-3mm}
\end{figure}

\subsection{Setup and Implementation Details}
We evaluate our method and baseline policies on the 28 Standard tasks of \textsc{TacMind-Bench} using the GelSight Mini sensor.
We run 20 episodes per task in the InD setting and in each of 12 OOD perturbation settings, yielding 260 episodes per task and 7,280 evaluation rollouts per model across the Precision, Recognition, and Dynamic suites.
All downstream policies are trained on the same 100 InD demonstrations per task using 8 NVIDIA A100 GPUs with an action horizon of $K=32$ and a tactile context length of $H=8$. We compare our method with many baselines: ACT+Tactile~\cite{univtac}, OpenPI0.5~\cite{pi05} and OpenDM0.5~\cite{opendm2026}, and N0-VTLA~\cite{team2026n_0}.
TacMind-0 is initialized directly from the pretrained OpenDM0.5 base policy and integrates our pretrained Tac-LeWM module.

\subsection{Comparison with Recent Works}
\textbf{TacMind-Bench Results.}
As shown in Table~\ref{tab:main_results}, TacMind-0 achieves the highest average success rates of 34.1\% InD and 20.2\% OOD.
Relative to its vision-only base policy OpenDM0.5, these results represent gains of 6.4 and 2.1 percentage points, respectively, with improvements across all three InD suites and most OOD settings.
Compared with OpenDM0.5, N0-VTLA yields lower InD success across all three suites and lower average OOD success.
TacMind-0 outperforms N0-VTLA across all reported settings, including Dynamic InD success of 19.1\% compared with 3.6\% for N0-VTLA.
The representative rollout in Fig.~\ref{fig:case1} illustrates the corresponding behavior: TacMind-0 responds to incipient rotational slip by adjusting its trajectory and inserting a bulb into a moving socket, whereas N0-VTLA fails.

\noindent\textbf{Real-world Results.} We also evaluate our method with OpenPI0.5 and N0-VTLA on multiple tasks, containing precise contact and tactile recognition tasks in the real-world robot arm. More details in Appendix.

\begin{table}[t]
    \centering
    \small
    \caption{Comparison of end-to-end (E2E) and post-aligned (PA) manipulation success rates (\%) across all the long-horizon precision and dynamic tasks in TacMind-Bench.}
    \vspace{-2mm}
    \label{tab:e2e_postalign_targets}
    \setlength{\tabcolsep}{14pt}
    \begin{tabular}{l*{7}{c}}
        \toprule
        \multirow{2}{*}{Model}
        & \multicolumn{2}{c}{Precision}
        & \multicolumn{2}{c}{Dynamic}
        & \multicolumn{2}{c}{Avg}
        & \multirow{2}{*}{$\Delta$ Avg (pp)} \\
        \cmidrule(lr){2-3}
        \cmidrule(lr){4-5}
        \cmidrule(lr){6-7}
        & E2E & PA
        & E2E & PA
        & E2E & PA & \\
        \midrule

        OpenPi0.5
        & 1.3 & 31.3
        & 2.5 & 27.5
        & 2.1 & 28.8
        & +26.7 \\

        OpenDM
        & 2.5 & 27.5
        & 8.1 & 25.6
        & 6.3 & 26.3
        & +20.0 \\

        TacMind-0
        & 3.8 & 33.8
        & 8.8 & 35.0
        & 7.1 & 34.6
        & +27.5 \\

        \bottomrule
    \end{tabular}%
    \vspace{-2mm}
\end{table}

\begin{table}[t]
\centering
\small
\caption{\textbf{Ablation study of TacMind-0.} Tac-FRS denotes predictive tactile reflection via flow-reversal steering.}
\vspace{-2mm}
\label{tab:ablation_tacmind0}
\setlength{\tabcolsep}{10pt}
\begin{tabular}{cc*{8}{c}}
        \toprule
        \multirow{2}{*}{Tactile}
        & \multirow{2}{*}{Tac-FRS}
        & \multicolumn{2}{c}{Precision}
        & \multicolumn{2}{c}{Recognition}
        & \multicolumn{2}{c}{Dynamic}
        & \multicolumn{2}{c}{Avg} \\
        \cmidrule(lr){3-4}
        \cmidrule(lr){5-6}
        \cmidrule(lr){7-8}
        \cmidrule(lr){9-10}
        & & InD & OOD
        & InD & OOD
        & InD & OOD
        & InD & OOD \\
        \midrule

        -- & --
        & 10.8 & 5.8
        & 48.2 & 36.1
        & 16.4 & 6.9
        & 27.7 & 18.1 \\

        $\checkmark$ & --
        & 17.5 & 6.7
        & 53.6 & 36.8
        & 16.4 & 7.1
        & 31.3 & 18.7 \\

        -- & $\checkmark$
        & 15.8 & 6.3
        & 47.7 & 35.6
        & 17.3 & 8.0
        & 28.9 & 18.5 \\

        $\checkmark$ & $\checkmark$
        & \textbf{20.8} & \textbf{8.1}
        & \textbf{56.4} & \textbf{38.0}
        & \textbf{19.1} & \textbf{9.0}
        & \textbf{34.1} & \textbf{20.2} \\

        \bottomrule
    \end{tabular}%
\vspace{-4mm}
\end{table}

\begin{table}[th]
\centering
\small
\caption{\textbf{Linear probing of frozen Tac-LeWM latent states.} Lower standardized error $\mathrm{MSE}_z$ and higher Pearson correlation $r$ indicate better representation quality.}
\vspace{-2mm}
\label{tab:taclewm_probe}
\setlength{\tabcolsep}{18pt}
\begin{tabular}{l cccc}
\toprule
Metric & Proprio & Gripper & Action & Pad Energy \\
\midrule
$\mathrm{MSE}_z \downarrow$ & 0.1816 & 0.1958 & 0.1841 & 0.2771 \\
Pearson $r \uparrow$        & 0.8981 & 0.9053 & 0.8976 & 0.7968 \\
\bottomrule
\end{tabular}
\end{table}

\begin{table}[th]
\centering
\caption{\textbf{Ablation study of Tac-LeWM architectural components} on \textsc{TacMind-Bench} .}
\vspace{-3mm}
\label{tab:taclewm_ablation_recon}
\small
\setlength{\tabcolsep}{14pt}
\begin{tabular}{lccccc}
\toprule
Variant & PSNR$\uparrow$ & MSE$\downarrow$ & RMSE$\downarrow$ & MAE$\downarrow$ & SSIM$\uparrow$ \\
\midrule
\multicolumn{6}{l}{\textit{Current Reconstruction}} \\
Full            & 31.83 & \textbf{0.00157} & \textbf{0.0341} & \textbf{0.0223} & \textbf{0.8748} \\
w/o Camera      & \textbf{32.66} & 0.00177 & 0.0348 & 0.0230 & 0.8649 \\
w/o Causal Attn & 31.45 & 0.00208 & 0.0385 & 0.0252 & 0.8451 \\
w/o CTR         & 30.85 & 0.00190 & 0.0372 & 0.0242 & 0.8620 \\
\midrule
\multicolumn{6}{l}{\textit{Future Reconstruction}} \\
Full            & \textbf{28.98} & \textbf{0.00175} & \textbf{0.0391} & \textbf{0.0263} & \textbf{0.8593} \\
w/o Camera      & 26.52 & 0.00255 & 0.0489 & 0.0342 & 0.8034 \\
w/o Causal Attn & 23.07 & 0.00558 & 0.0724 & 0.0494 & 0.7116 \\
w/o CTR         & 23.42 & 0.00567 & 0.0714 & 0.0518 & 0.7088 \\
\bottomrule
\end{tabular}

\end{table}

 \begin{figure}[th]
   \centering
   \begin{minipage}[t]{0.49\columnwidth}
     \centering
     \includegraphics[width=\linewidth]{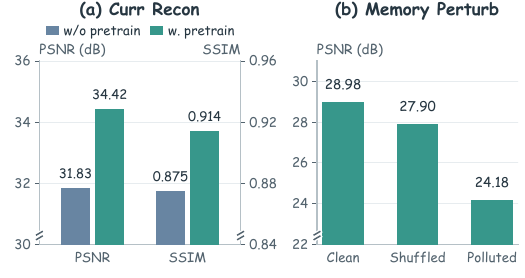}
     \vspace{-8mm}
     \caption{\textbf{Reconstruction and memory perturbation analysis.}
     \textbf{(a)} Effect of multi-sensor pretraining on current-frame tactile
     reconstruction.
     \textbf{(b)} Future tactile prediction PSNR under clean, shuffled, and
     polluted memory settings.}
     \label{fig:ab_bars}
   \end{minipage}
   \hfill
   \begin{minipage}[t]{0.49\columnwidth}
     \centering
     \includegraphics[width=\linewidth]{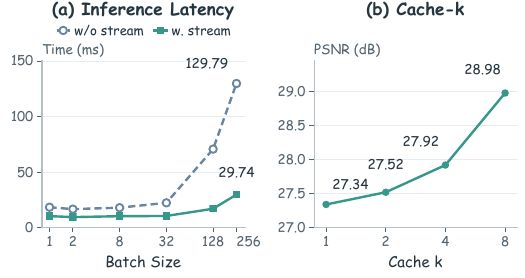}
     \vspace{-8mm}
     \caption{\textbf{Streaming latency and temporal horizon analysis.}
     \textbf{(a)} Tactile encoder inference latency across batch sizes with and
     without streaming KV caching.
     \textbf{(b)} Future tactile prediction PSNR across cache lengths $k$.}
     \label{fig:ab_lines}
   \end{minipage}
   \vspace{-3mm}
\end{figure}
   
\subsection{Ablation Study}
\noindent\textbf{End-to-End vs. Post-Aligned Manipulation.}
Prior contact-rich manipulation benchmarks, such as UniVTac~\cite{univtac}, primarily evaluate policies from post-aligned states, overlooking contact interactions during the preceding pick-and-transport stage. However, these interactions introduce additional challenges and can significantly affect overall task success. By incorporating the complete manipulation process, our benchmark provides a more challenging and comprehensive evaluation of contact-rich manipulation capabilities. To quantify the impact of the preceding stages, we further compare end-to-end and post-aligned success rates on the end-to-end tasks in TacMind-Bench Standard, \textit{i.e.}, Precision(Task0-3) and Dynamic (Task 3-10).

\noindent\textbf{TacMind-0 Mechanisms.}
Streaming tactile conditioning supplies the larger InD gain, while predictive tactile reflection further improves every suite as shown in Table~\ref{tab:ablation_tacmind0}.
Conditioning on $z_t^\tau$ raises average InD success from 27.7\% to 31.3\%, with the largest gains in Precision and Recognition, where physical feedback conveys contact geometry and object properties.
Adding reflection further increases average InD success from 31.3\% to 34.1\% and average OOD success from 18.7\% to 20.2\%, including an OOD Precision improvement from 6.7\% to 8.1\%.
This benefit aligns with the role of reflection: when viewpoint or scale shifts introduce alignment offsets in $A_t^{(0)}$, it predicts the contact consequences and uses flow-reversal steering to refine the proposed action.

\noindent\textbf{Tactile Representation Quality.}
Multi-sensor pretraining improves both PSNR and SSIM for tactile reconstruction (Fig.~\ref{fig:ab_bars}(a)), and frozen Tac-LeWM states retain information about robot kinematics and tactile deformation as shown in Table~\ref{tab:taclewm_probe}.
Linear probes achieve strong correlations with proprioception, gripper state, actions, and tactile pad deformation energy.
To examine the architectural contributions, Table~\ref{tab:taclewm_ablation_recon} compares three variants trained on \textsc{TacMind-Bench} Standard: w/o Camera removes auxiliary visual conditioning during training, w/o Causal Attn replaces temporal causal attention with single-frame self-attention, and w/o CTR replaces CTR with standard ViT raster positional encoding.
Current-frame reconstruction remains comparable across these variants, whereas removing causal attention or CTR sharply worsens future reconstruction on every metric and increases pad-energy probing error as shown in Table~\ref{tab:taclewm_ablation_recon}.
The two components thus contribute most clearly to forecasting contact dynamics and representing contact deformation.

\noindent\textbf{Temporal Context and Streaming Efficiency.}
Tac-LeWM benefits from ordered contact history, while streaming KV caching reduces the cost of retaining it.
Extending the cache horizon from $k=1$ to $k=8$ steadily improves future tactile prediction, as shown in Fig.~\ref{fig:ab_lines}(b); shuffling token order or polluting the cache with irrelevant frames lowers reconstruction accuracy, as shown in Fig.~\ref{fig:ab_bars}(b).
These interventions show that prediction depends on temporal contact context and its ordering.
Streaming KV caching substantially lowers per-step encoding latency both for single-stream robot inference and at larger batch sizes used in parallel simulation, as shown in Fig.~\ref{fig:ab_lines}(a).
Combined with the compiled three-pass action expert in Appendix, this reuse supports fast closed-loop predictive reflection.

\raggedbottom
\section{Conclusion}

We introduced \textsc{TacMind-Bench} to evaluate tactile manipulation across precise contact, recognition under ambiguity, and dynamic interaction, with 28 Standard evaluation tasks, 100 Extra demonstration tasks, and 12 controlled perturbations for testing OOD robustness in simulation.
We also introduce Tac-LeWM, which learns streaming tactile interaction states through action-conditioned latent prediction, CTR positional encoding, and causal KV caching.
TacMind-0 uses these states to condition action generation and then refines its actions through predictive tactile reflection with flow-reversal steering.
Experiments demonstrate substantial gains in multi-task manipulation and OOD robustness, while the ablations establish complementary contributions from current tactile conditioning and predictive tactile reflection.

\newpage

\subsection*{AI use statement}

In this work, we used generative AI tools to assist with polishing and improving the clarity of the manuscript, debugging code, and supporting data collection for our benchmark. We did not use generative AI tools to independently formulate the main research claims or conclusions. All AI-assisted manuscript edits were reviewed by the authors. AI-assisted code debugging was followed by manual inspection and testing to verify correctness. For benchmark data collection, AI-generated or AI-assisted outputs were reviewed and filtered by the authors according to our data collection protocol and quality-control criteria. We take full responsibility for the final content of this work, including all text, claims, code, and benchmark data produced with the aid of generative AI.

\subsection*{Ethics statement}

Our benchmark is built in the simulation. The collected data will be released, but there is no human subjects data recorded.

\subsection*{Reproducibility statement}

The design of our benchmark is introduced in Section 3. 
The framework and training/inference setup of our model are introduced in Section 4.
The details of the benchmark and experiment setup are stated in Section 5 and the appendix.
The data and code of the benchmark and the model will be made public.

\bibliography{iclr2027_conference}
\bibliographystyle{iclr2027_conference}

\clearpage
\appendix
\section{Benchmark Data Collection}

\subsection{Agentic Manipulation Data Generation}

\noindent\textbf{Motivation.}
Agentic data generation is particularly suitable for collecting demonstrations in contact-rich tasks, especially for the precise contact regime.
We initially followed prior work~\cite{libero} and used a SpaceMouse for human teleoperation.
However, due to our strict seated completion criteria and the limited precision of manual teleoperation, this approach proved inefficient and frequently produced jittery, suboptimal trajectories.
Our structured agentic pipeline addresses this collection bottleneck by generating smooth, precise trajectories through reusable task controllers.

\noindent\textbf{Settings.}
We use a large-language-model coding agent to synthesize and debug closed-loop action experts in Isaac Sim, with a harness that supports iterative rollout and diagnosis.
We formulate agentic manipulation as a modular controller program with an explicit closed-loop structure that makes decisions based on visual observations, robot proprioception, and tactile feedback.
The control architecture decouples free-space motion from contact-rich manipulation.
Motion planning handles approaching and transporting objects in free space, whereas local closed-loop control executes grasping, alignment, insertion, pressing, fitting, and fastening.
Tactile or geometric conditions rather than fixed time steps trigger transitions between execution stages.

\noindent\textbf{Environment and Sensors.}
We use Isaac Sim 5.1~\cite{NVIDIA_Isaac_Sim} as the simulation backend and CuRobo~\cite{curobo} for manipulation trajectory planning.
We integrate two vision-based optical tactile sensors, namely GelSight Mini and Daimon (DM-Tac), both of which infer contact states from camera observations of elastomer deformation.
GelSight Mini provides high-resolution sensing of local contact geometry and surface texture, whereas Daimon emphasizes dense dynamic perception of contact shape, normal pressure, tangential shear, and slip.
The robot platform is a Franka Emika arm instantiated with two distinct gripper configurations equipped with GelSight Mini and Daimon sensors, respectively.
Because the two sensor-equipped grippers differ in physical geometry, we tune the motion trajectories and data-collection experts separately for each configuration.

\noindent\textbf{Success Criteria.}
A task is considered successful only when the manipulated object reaches a stable seated state, such as full insertion, complete press fitting, or completion of a target rotational angle.
In discrimination and search tasks, decisions must be grounded in tactile observations acquired through physical interaction.
To ensure high demonstration quality, the success criteria enforced during data collection are stricter than those used during policy evaluation.

During agentic data collection, trajectories judged successful online do not always succeed when replayed open-loop under the identical scene configuration.
During collection, the controller replans and adjusts actions in response to small physical variations. Open-loop replay re-executes the recorded commands without these corrections and is therefore sensitive to minor simulator nondeterminism.
This sensitivity is particularly pronounced in fine-grained manipulation tasks where millimeter-scale deviations lead to failure.
Therefore, we admit only trajectories that also succeed during deterministic open-loop replay as valid training demonstrations.

\noindent\textbf{Skills.}
We organize the agent's capabilities into modular skills for asset processing, task generation, environment construction, and trajectory synthesis. Each skill supports a stage of benchmark construction, as summarized in Table~\ref{tab:tacbench_skills}.

\begin{table*}[t]
\centering
\caption{Summary of data-collection skills.}
\label{tab:tacbench_skills}
\small
\begin{tabularx}{\linewidth}{
    >{\raggedright\arraybackslash}p{0.27\linewidth}
    >{\raggedright\arraybackslash}X
}
\toprule
\textbf{Skill} & \textbf{Functional Scope} \\
\midrule
\texttt{expert-trajectory} &
Development and refinement of expert policies for our benchmark, including tactile grasping, CuRobo-based free-space motion planning, insertion and placement, success-condition specification, and trajectory-smoothness optimization. \\
\texttt{dynamic-collect} &
Collection and same-environment replay-gated evaluation of dynamic conveyor-belt tasks \texttt{tac\_dynamic}, together with systematic diagnosis of execution hangs and zero-episode dataset failures. \\
\texttt{reset-randomization} &
Design and validation of reset-time perturbations for the benchmark suite, including \texttt{pose\_range}, object permutation, and \texttt{layout\_xy}, with consistency verification between specification files and runtime perturbation bounds. \\
\texttt{generate-tactile-task} &
Generation and validation of collectible tactile manipulation environments from natural-language task specifications. \\
\texttt{object-expert-repair} &
Diagnosis and remediation of geometry and collision issues, insertion failures, false-positive success conditions, and expert trajectory synthesis consistent with teleoperated behavior. \\
\bottomrule
\end{tabularx}
\end{table*}

\noindent\textbf{Built-in CLI.}
The agent interacts with the simulator dynamically to diagnose physical failures during rollout.
Built-in command-line functions expose simulation logs, capture rendered observations, and process collected trajectories. These tools support expert generation and diagnosis of environment or execution errors (Table~\ref{tab:builtin_functions}).

\begin{table*}[t]
\centering
\caption{Built-in functions and their roles in the data-processing pipeline.}
\label{tab:builtin_functions}
\small
\begin{tabularx}{\linewidth}{
    >{\raggedright\arraybackslash}p{0.20\linewidth}
    >{\raggedright\arraybackslash}X
}
\toprule
\textbf{Function} & \textbf{Role in the Data-Processing Pipeline} \\
\midrule
\texttt{collect} &
Performs data collection, validation, and persistence while exposing diagnostic information and runtime task traces to support agent-driven analysis of collection failures and their underlying patterns. \\
\texttt{expert} &
Dispatches the task-specific expert procedure to generate demonstrations for the target manipulation task. \\
\texttt{replay\_gate} &
Replays collected trajectories in the corresponding environment and evaluates execution correctness as a quality-control gate for dataset admission. \\
\texttt{capture} &
Captures visual renders from arbitrary tasks in Isaac Sim for inspection, debugging, and downstream multimodal processing. \\
\texttt{validate\_dataset} &
Inspects dataset metadata and verifies individual episodes against the predefined episode schema to ensure structural consistency. \\
\texttt{convert\_dataset} &
Converts datasets across storage formats, processes action sequences, and filters redundant frames to produce training-ready datasets. \\
\bottomrule
\end{tabularx}
\end{table*}

\noindent\textbf{Human in the Loop.}
Constructing simulation environments, executing contact-rich tactile tasks, and synthesizing trajectory-generation experts solely through an autonomous coding agent remain challenging.
Due to limitations in three-dimensional spatial reasoning, physical dynamics understanding, and rollout video interpretation, automated agents often struggle to diagnose subtle failure modes and repair contact parameters autonomously.
We therefore incorporate human verification throughout the pipeline to check simulation assets, task environments, and training demonstrations.

\noindent\textbf{Touch in Data Collection.}
Beyond synchronously recording tactile RGB images and marker displacement fields, tactile feedback is actively used within the agentic controller during data generation.
During grasping and contact verification, we feed the indentation depth measured by the tactile sensors into an adaptive gripper controller that regulates gripper width to maintain grasping force within a safe target range.
For sorting and recognition, simulator state cues schedule candidate objects, while rule-based procedures govern sequential probing. Identification decisions are made only after physical tactile contact.
For tactile search tasks, the controller determines whether the target object has been located directly from contact responses registered by the tactile sensors.

\noindent\textbf{Randomness in Data Collection.}
We randomize object locations and the order of tactile probing, particularly in recognition tasks, to vary the spatial and sequential context in which the robot explores and identifies objects.

\noindent\textbf{Action Representation.}
We record and execute 8-DoF absolute joint actions comprising seven arm joint angles and one continuous gripper command.
In preliminary experiments, replaying recorded trajectories in end-effector pose space yielded low success rates because inverse kinematics solver errors degraded contact precision and trajectory quality.

\noindent\textbf{Additional Modalities.}
Isaac Sim also supports depth maps, semantic segmentation masks, object poses, and tracking states. We provide extensible interfaces for exporting these additional modalities.

\subsection{Data Collection Principles and Discussion}

\noindent\textbf{Trajectory Length.}
For demonstrations of a given task, synthesized trajectories are designed to minimize redundant motion and unnecessary corrective loops.
Excessive trajectory length increases the effective task horizon and exacerbates compounding errors in imitation learning~\cite{ross2011reduction, shi23b, zhao2023learning}.
Prior work also identifies path length and trajectory compactness as key indicators of demonstration quality~\cite{sakr2025consistency}.
Our agentic experts therefore execute tasks concisely while preserving necessary safety clearances and contact transitions.

\noindent\textbf{Trajectory Speed.}
Demonstrations should exhibit smooth and temporally consistent velocity profiles rather than abrupt speed changes.
Sudden accelerations, frequent stops, and oscillatory corrections introduce action variance that hinders policy convergence.
Motion smoothness characterized by low jerk is closely associated with high-quality visuomotor learning~\cite{flash1985coordination,sakr2025consistency}.
Our trajectory synthesis pipeline therefore maintains steady motion in free space and smoothly decelerates during contact establishment and precision alignment.

\noindent\textbf{Trajectory Consistency.}
Demonstrations within the same task follow consistent motion strategies and execution patterns.
Large variations in trajectory geometry across demonstrations increase multimodality and ambiguity in the target action distribution.
Prior studies confirm that structural consistency across demonstrations correlates strongly with downstream task success and generalization~\cite{sakr2025consistency}.

\noindent\textbf{Action Predictability.}
Our expert controllers avoid arbitrary variations in action selection under similar task states.
Deterministic state-to-action mappings reduce unnecessary action ambiguity and improve the predictability of demonstrations, which serves as a core indicator of robot dataset quality~\cite{shi2026diversity}.

\subsection{Task Suites and Benchmark Details}
\label{app:tasks}

\begin{figure*}[t]
\centering
\includegraphics[width=\linewidth]{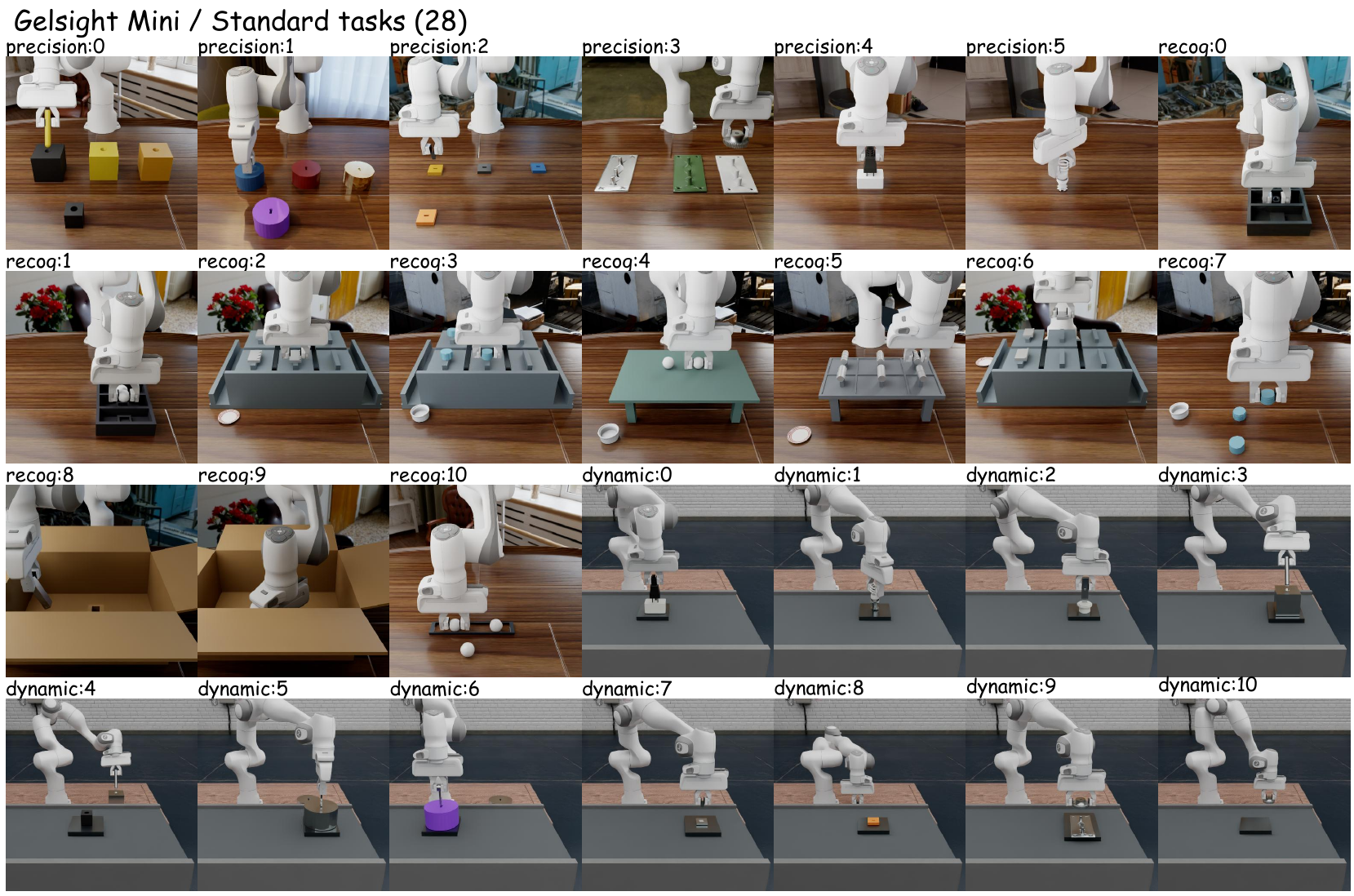}
\caption{Overview of the 28 Standard tasks in \textsc{TacMind-Bench} with GelSight Mini.}
\label{fig:task_examples}
\end{figure*}

\begin{figure*}[t]
\centering
\includegraphics[width=\linewidth]{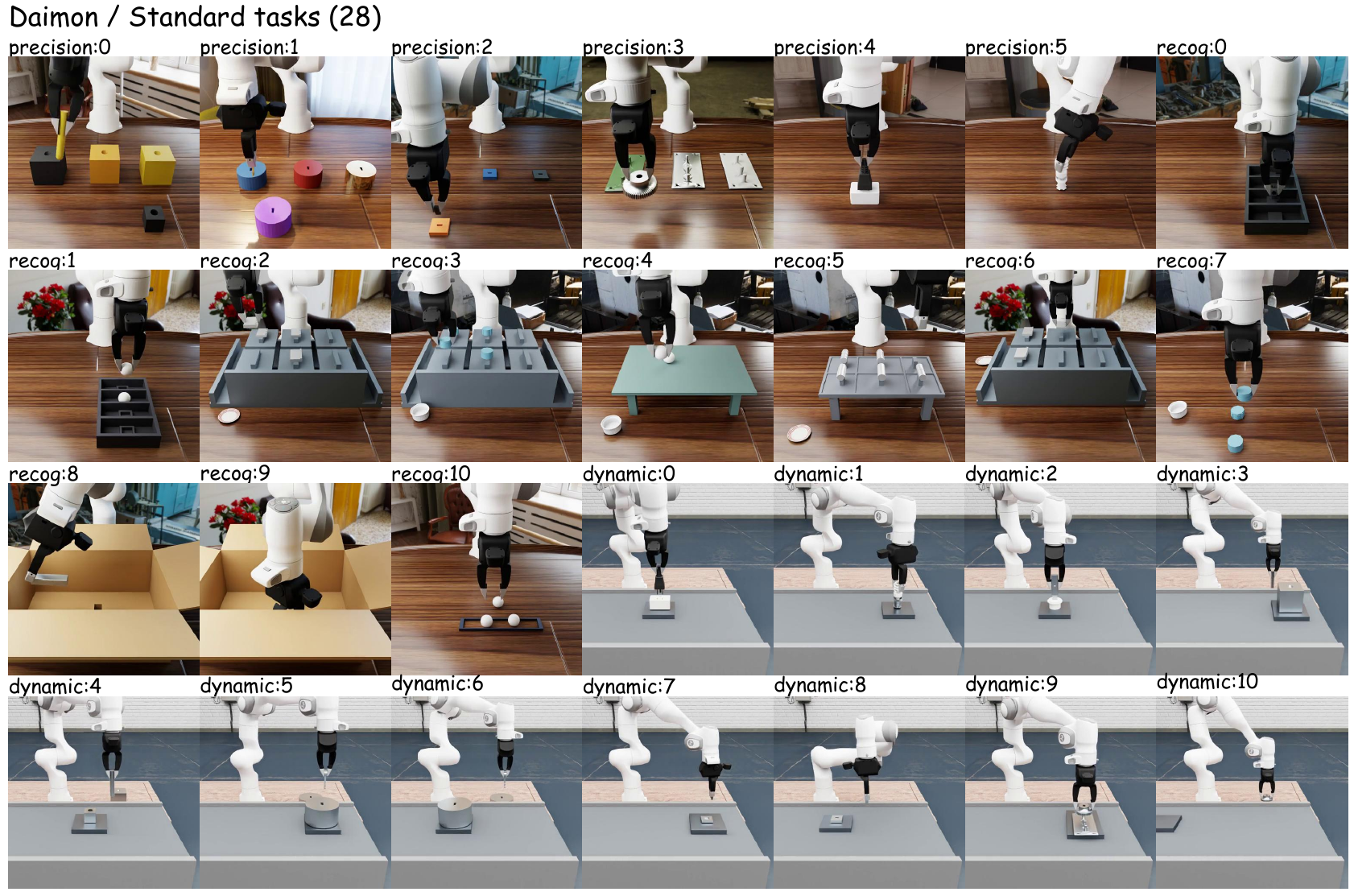}
\caption{Overview of the 28 Standard tasks in \textsc{TacMind-Bench} with Daimon.}
\label{fig:task_examples_2}
\end{figure*}

\begin{figure*}[t]
\centering
\includegraphics[width=\linewidth]{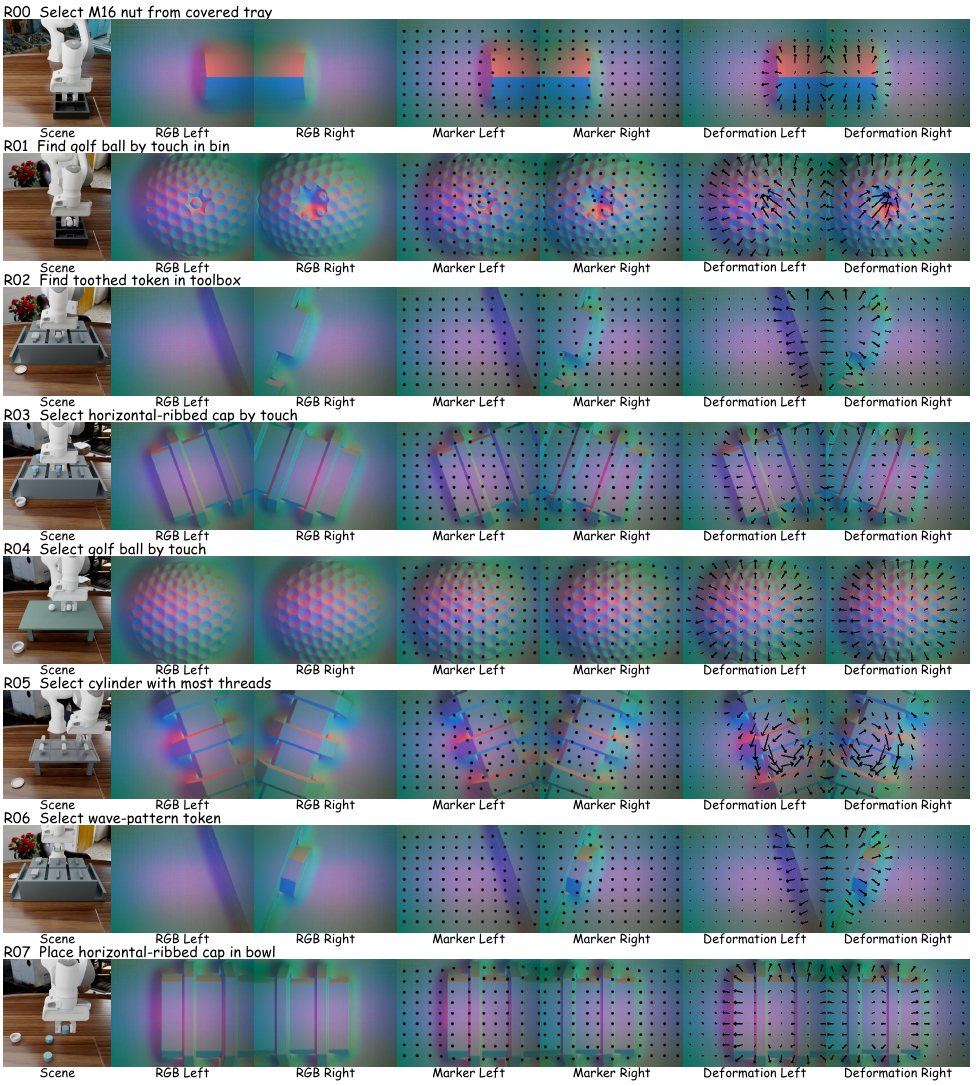}
\caption{Visualization of tactile observations from GelSight Mini.}
\label{fig:gsmini_tactile_vis}
\end{figure*}

\begin{figure*}[t]
\centering
\includegraphics[width=\linewidth]{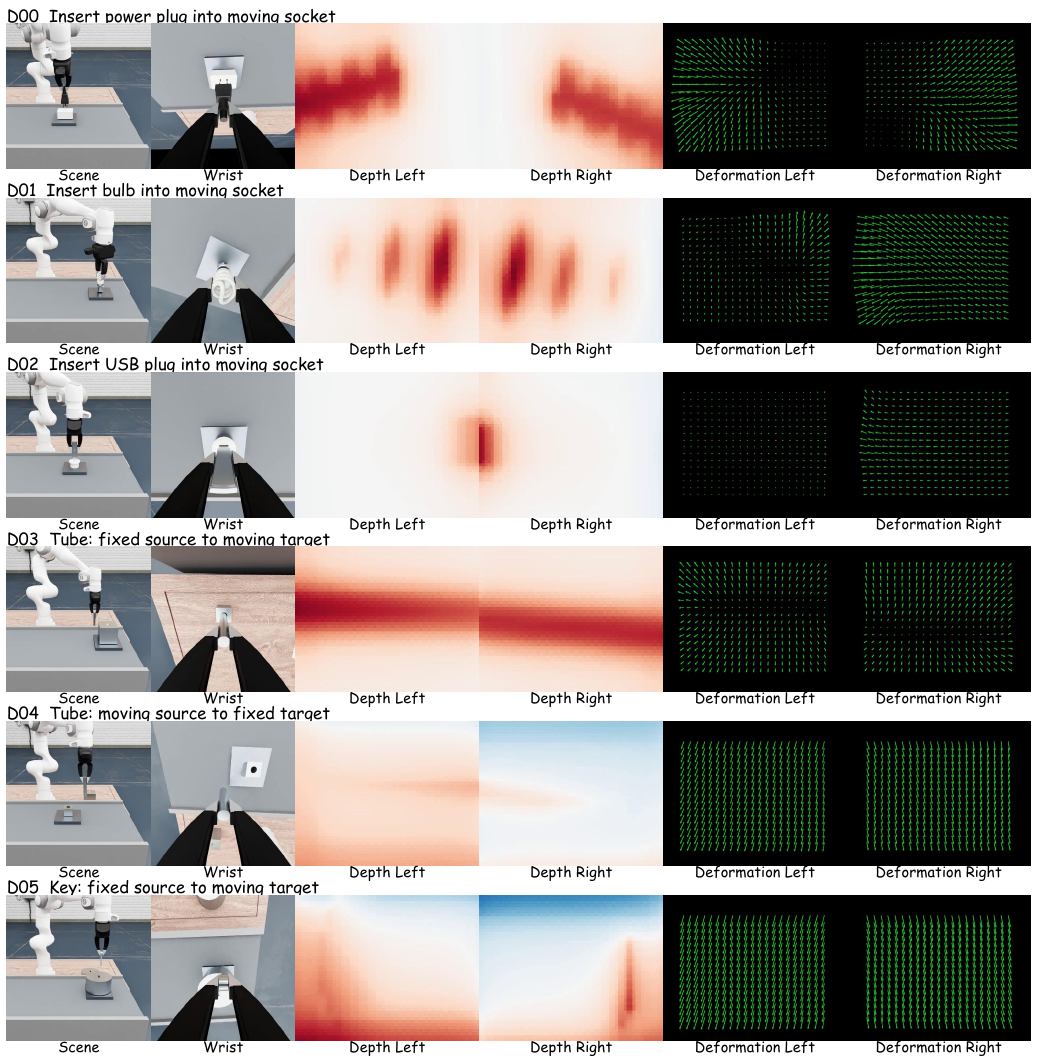}
\caption{Visualization of tactile observations from Daimon.}
\label{fig:daimon_tactile_vis}
\end{figure*}

The Standard suite comprises 28 tasks spanning the three evaluation regimes, namely precise contact, recognition under ambiguity, and dynamic interaction.
All Standard tasks share a unified observation interface instantiated across two sensor-equipped gripper embodiments, GelSight Mini and Daimon, and each task provides 100 expert demonstrations collected via our agentic pipeline.
Figure~\ref{fig:task_examples} and Figure~\ref{fig:task_examples_2} illustrate representative scene observations across the 28 Standard tasks for the GelSight Mini and Daimon configurations, while Figure~\ref{fig:gsmini_tactile_vis} and Figure~\ref{fig:daimon_tactile_vis} visualize the representative tactile image and marker signals during contact.
The complete list of natural-language instructions for the Standard suite is provided in Table~\ref{tab:tasks_all}.

\noindent\textbf{Precise Contact Regime.}
Comprising 6 tasks, the precise contact regime evaluates tabletop contact regulation in static scenes, including picking and inserting a tube, key, connector, or gear into a corresponding hole base, lock, or port, as well as inserting a three-prong power plug and a light bulb into their respective sockets.

\noindent\textbf{Recognition under Ambiguity Regime.}
The 11 recognition tasks require tactile cues, including texture, rib orientation, edge contour, and size, to identify visually ambiguous or occluded targets. Seven tasks conceal objects inside boxes or covered trays, while four present visually indistinguishable objects in open view.

\noindent\textbf{Dynamic Interaction Regime.}
The 11 dynamic tasks evaluate insertion and press-fitting under externally driven motion. A translating conveyor carries either the target fixture or the initial object base, so alignment must adapt to continuous relative motion.

\begin{figure*}[t]
\centering
\includegraphics[width=\linewidth]{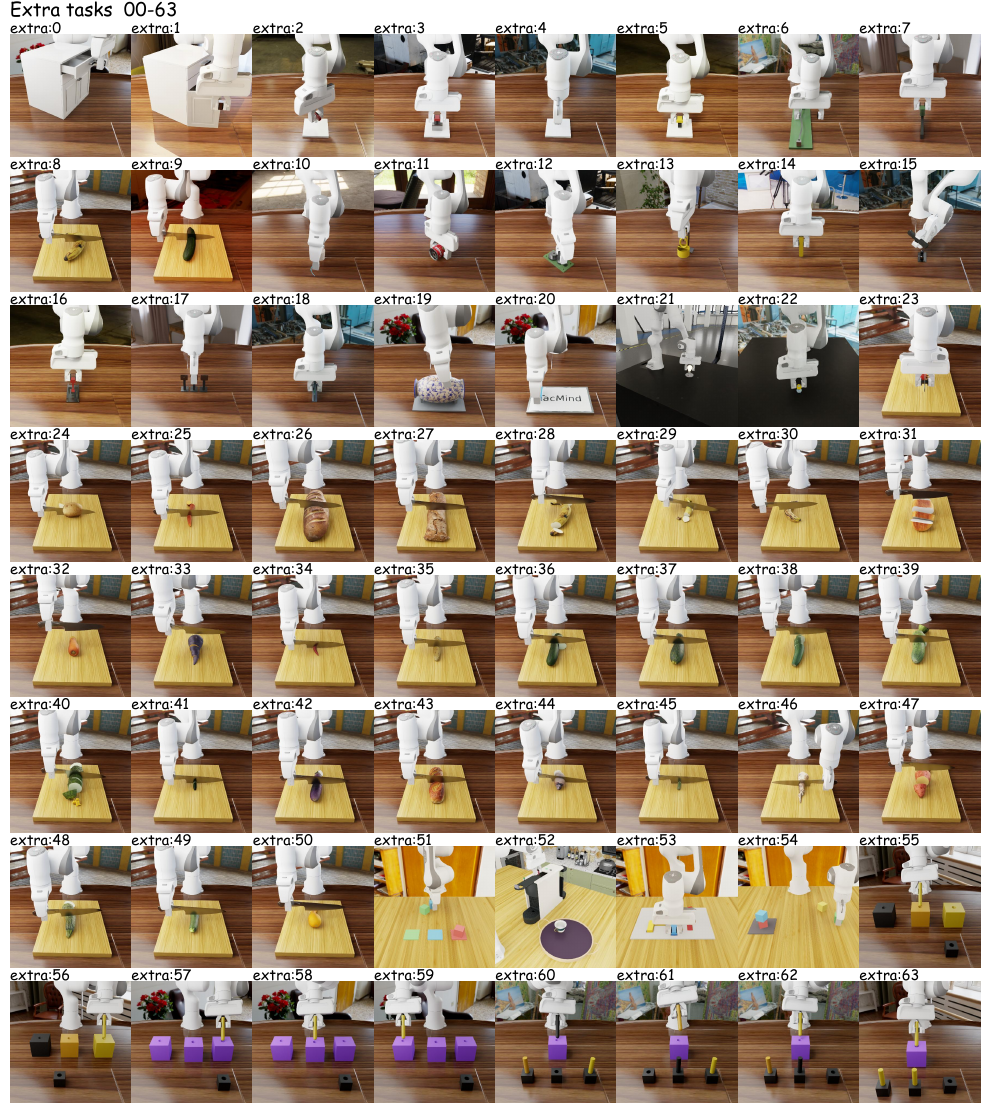}
\caption{Visualization of Extra suite tasks 00 to 63 in \textsc{TacMind-Bench}.}
\label{fig:extra_task_0_63}
\end{figure*}

\begin{figure*}[t]
\centering
\includegraphics[width=\linewidth]{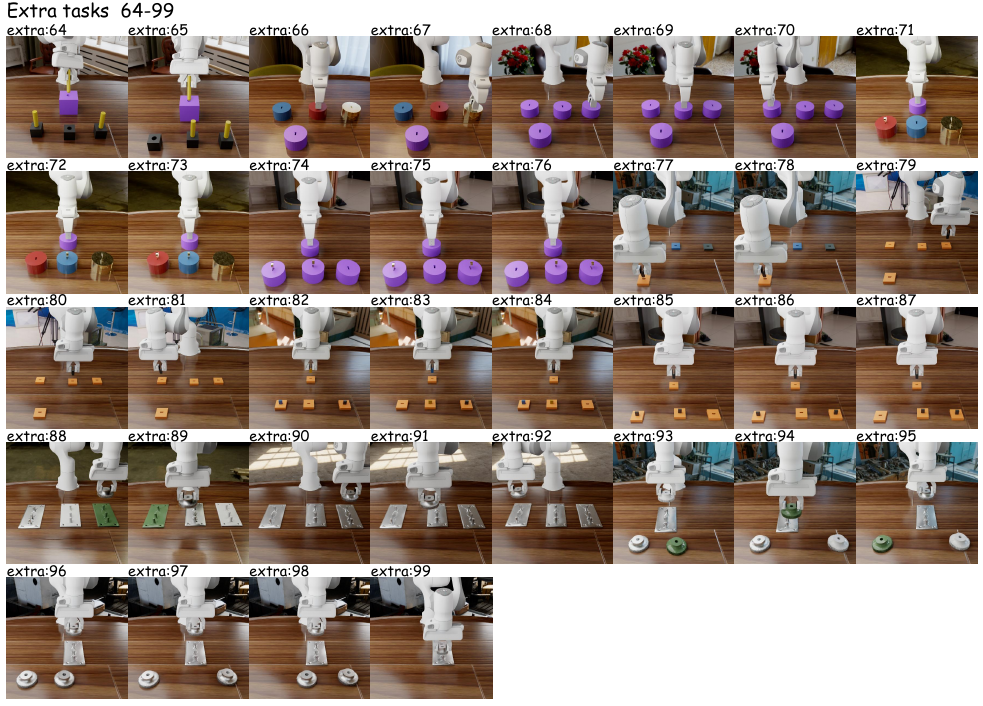}
\caption{Visualization of Extra suite tasks 64 to 99 in \textsc{TacMind-Bench}.}
\label{fig:extra_task_64_99}
\end{figure*}

\noindent\textbf{Extra Suite.}
The Extra suite contains 100 demonstration tasks covering 11 contact behaviors, comprising 27 articulated-mechanism, non-prehensile, and surface-interaction tasks such as opening drawers and doors, pressing buttons, sliding latches, rolling cans, and wiping surfaces, alongside 30 tool-based food cutting tasks and 43 compositional assembly tasks, as visualized in Figure~\ref{fig:extra_task_0_63} and Figure~\ref{fig:extra_task_64_99}.
The complete catalog of task identifiers and natural-language instructions is detailed in Table~\ref{tab:tasks_extra}, where tasks sharing similar instructions correspond to distinct object mesh geometries, cutting-plane counts, or spatial configurations.

\noindent\textbf{Evaluation Settings.}
Each Standard task is evaluated across 1 InD setting and 12 OOD perturbation settings comprising 4 illumination conditions, 6 camera viewpoints, and 2 object scales.
We conduct 20 evaluation trials per setting, yielding 260 trials per task and 7,280 rollouts across all 28 Standard tasks.

\begin{table*}[t!]
\centering
\caption{All 28 \textsc{TacMind-Bench} Standard tasks grouped by evaluation regime. In the Dynamic regime, either the target fixture or the source base is mounted on a moving conveyor platform.}
\label{tab:tasks_all}
\footnotesize
\begin{tabularx}{\linewidth}{
    >{\raggedright\arraybackslash}p{0.08\linewidth}
    >{\raggedright\arraybackslash}p{0.04\linewidth}
    >{\raggedright\arraybackslash}X
}
\toprule
\textbf{Regime} & \textbf{Task} & \textbf{Language Instruction} \\
\midrule
\multirow{6}{*}{Precise}
 & 00 & Pick up the tube and insert it into the black rubber hole base. \\
 & 01 & Pick up the key and insert it into the blue plastic lock. \\
 & 02 & Pick up the connector and insert it into the orange port. \\
 & 03 & Pick up the gear and insert it into the white painted base. \\
 & 04 & Insert the three-prong power plug into the power socket. \\
 & 05 & Insert the bulb into the bulb socket. \\
\midrule
\multirow{11}{*}{Recog}
 & 00 & Pick the larger M16 nut instead of the M8 nut from the covered tray, judging size by touch. \\
 & 01 & Retrieve the dimpled golf ball rather than the smooth ball from the covered bin using touch. \\
 & 02 & Explore the covered compartment tray by touch, find the service token with the sawtooth edge, and place it on the rubber service mat. \\
 & 03 & Among the caps in the covered recycling tray, find the cap with horizontal ribs rather than vertical ribs and place it in the recycle cup. \\
 & 04 & Explore the two balls by touch, identify the dimpled golf ball rather than the smooth table-tennis ball, and place it in the left cup. \\
 & 05 & Among the six equal-size cylinders in the open parts box, find the one with the most circumferential threads and place it on the inspection tray. \\
 & 06 & Among the tokens in the covered compartment tray, find the one with the wave-shaped edge and place it on the rubber service mat. \\
 & 07 & Among the three caps on the table, use touch to find the cap with horizontal ribs and place it in the bowl. \\
 & 08 & Pick up the horizontal rectangular peg from the box bottom, lift it vertically, and insert it. \\
 & 09 & Probe inside the occluded box to locate and retrieve the concealed target object using tactile feedback. \\
 & 10 & Distinguish the golf ball from the table tennis ball by touch and lift the golf ball. \\
\midrule
\multirow{11}{*}{Dynamic}
 & 00 & Insert the three-prong power plug already held by the gripper into the power socket on the moving conveyor platform. \\
 & 01 & Insert the bulb already held by the gripper into the bulb socket on the moving conveyor platform. \\
 & 02 & Insert the USB plug already held by the gripper into the USB socket on the moving conveyor platform. \\
 & 03 & Pick the test tube from the table and insert it into the angled hole base mounted on the conveyor platform. \\
 & 04 & Pick the test tube from the moving source base and insert it into the angled hole base on the table. \\
 & 05 & Pick the key from the table and insert it into the target lock mounted on the conveyor platform. \\
 & 06 & Pick the key from the moving source lock and insert it into the target lock on the table. \\
 & 07 & Pick the HDMI connector from the table and insert it into the HDMI slot mounted on the conveyor platform. \\
 & 08 & Pick the HDMI connector from the moving source platform and insert it into the HDMI slot on the table. \\
 & 09 & Pick the gear from the table and press it into the gear base mounted on the conveyor platform. \\
 & 10 & Pick the gear from the moving source platform and press it into the gear base on the table. \\
\bottomrule
\end{tabularx}
\end{table*}

\begin{table*}[p]
\centering
\vspace{-14mm}
\caption{All 100 \textsc{TacMind-Bench} Extra tasks. Task IDs correspond to zero-based indices in the Extra suite catalog, where entries sharing base instructions correspond to distinct object meshes, cutting-plane counts, or spatial layouts.}
\vspace{-4mm}
\label{tab:tasks_extra}
\scriptsize
\setlength{\tabcolsep}{1pt}
\renewcommand{\arraystretch}{0.92}
\begin{tabularx}{\textwidth}{
    c >{\raggedright\arraybackslash}X |
    c >{\raggedright\arraybackslash}X
}
\toprule
\textbf{ID} & \textbf{Language Instruction} & \textbf{ID} & \textbf{Language Instruction} \\
\midrule
00 & Pull the drawer open by at least 12 cm. & 50 & Grasp the knife and cut the yellow zucchini at every scored plane. \\
01 & Grasp the handle and open the hinged door by at least 60 degrees. & 51 & Sort all blocks into the matching color zones. \\
02 & Turn the valve counterclockwise by 90 degrees relative to its initial angle. & 52 & Load the coffee machine, place the cup under the spout, and press the button to brew coffee into the cup. \\
03 & Press the spring button at least 8 mm for 0.3 seconds, then release it. & 53 & Push each shape into its matching recess without lifting it. \\
04 & Move the bistable lever from the negative well to the positive well, then let go. & 54 & Stack every block into one aligned, stable tower on the target pad. \\
05 & Pull the cap completely out of its spring-loaded retaining shoes and move it away by at least 3 cm. & 55 & Pick up the tube and insert it into the orange plastic hole base. \\
06 & Follow the thin sinusoidal rail from its left end to its right end while maintaining tactile contact. & 56 & Pick up the tube and insert it into the yellow plastic hole base. \\
07 & Pivot the block upright by 90 degrees while keeping it supported by the table and allowing rotation within the gripper. & 57 & Pick up the tube and insert it into the right hole base. \\
08 & Use the knife held in the gripper to cut the banana at its middle scored plane, and lift the blade clear while leaving both portions on the board. & 58 & Pick up the tube and insert it into the middle hole base. \\
09 & Use the knife held in the gripper to cut the cucumber at its middle scored plane, and lift the blade clear while leaving both portions on the board. & 59 & Pick up the tube and insert it into the left hole base. \\
10 & Push the block along the complete S-shaped path while maintaining tactile and tabletop contact. & 60 & Pick up the black rubber tube and insert it into the hole base. \\
11 & Roll the can through half a turn along the table without sliding or lifting it. & 61 & Pick up the orange plastic tube and insert it into the hole base. \\
12 & Turn the medium gear by 90 degrees so that both passive output gears rotate through their required angles. & 62 & Pick up the yellow plastic tube and insert it into the hole base. \\
13 & Lift the mug, thread its handle over the hook, lower it, and release it so that the hook supports its weight. & 63 & Pick up the right tube and insert it into the hole base. \\
14 & Lift the rod, hold the hand still, allow a 20 mm downward slip, then tighten the fingers to arrest it without regrasping. & 64 & Pick up the middle tube and insert it into the hole base. \\
15 & Pick up the horizontal rod near its center, orient it upright, place its flat end beside the cradle, then release and retreat while it stands unsupported. & 65 & Pick up the left tube and insert it into the hole base. \\
16 & Grasp the raised latch grip, depress its spring-loaded lug below the keeper, slide it clear along the rail, then release and retreat while the plunger returns. & 66 & Pick up the key and insert it into the red plastic lock. \\
17 & Briefly touch each of the three inspection coupons, lifting clear between probes, then indicate the raised defect with a steady final gel contact. & 67 & Pick up the key and insert it into the brass lock. \\
18 & Withdraw the RAM module at least 2 cm, reinsert its keyed edge, then release it and retreat. & 68 & Pick up the key and insert it into the right lock. \\
19 & Grasp the cloth-wrapped sponge, wipe the dirt from the floral porcelain vase secured in its cradle, then lift the sponge clear. & 69 & Pick up the key and insert it into the middle lock. \\
20 & Grasp the felt eraser, erase the TacMind lettering from the rectangular whiteboard, then lift the eraser clear. & 70 & Pick up the key and insert it into the left lock. \\
21 & Grasp the light bulb and screw it fully into the socket. & 71 & Pick up the key from the blue plastic source lock and insert it into the target lock. \\
22 & Grasp the nut and thread it fully onto the bolt. & 72 & Pick up the key from the red plastic source lock and insert it into the target lock. \\
23 & Grasp the knife and dice the tomato at every scored plane. & 73 & Pick up the key from the brass source lock and insert it into the target lock. \\
24 & Grasp the knife and dice the potato at every scored plane. & 74 & Pick up the right key and insert it into the lock. \\
25 & Grasp the knife and cut the carrot at every scored plane (mesh A). & 75 & Pick up the middle key and insert it into the lock. \\
26 & Grasp the knife and cut the baguette at every scored plane (mesh A). & 76 & Pick up the left key and insert it into the lock. \\
27 & Grasp the knife and cut the baguette at every scored plane (mesh B). & 77 & Pick up the connector and insert it into the blue port. \\
28 & Grasp the knife and cut the banana at every scored plane (mesh A). & 78 & Pick up the connector and insert it into the black port. \\
29 & Grasp the knife and cut the banana at every scored plane (mesh B). & 79 & Pick up the connector and insert it into the right port. \\
30 & Grasp the knife and cut the banana at every scored plane (mesh C). & 80 & Pick up the connector and insert it into the middle port. \\
31 & Grasp the knife and cut the bread loaf at every scored plane. & 81 & Pick up the connector and insert it into the left port. \\
32 & Grasp the knife and cut the carrot at every scored plane (mesh B). & 82 & Pick up the orange connector and insert it into the port. \\
33 & Grasp the knife and cut the purple carrot at every scored plane. & 83 & Pick up the blue connector and insert it into the port. \\
34 & Grasp the knife and cut the chili at every scored plane. & 84 & Pick up the black connector and insert it into the port. \\
35 & Grasp the knife and cut the corn at every scored plane. & 85 & Pick up the right connector and insert it into the port. \\
36 & Grasp the knife and cut the cucumber at every scored plane (mesh A). & 86 & Pick up the middle connector and insert it into the port. \\
37 & Grasp the knife and cut the cucumber at every scored plane (mesh B). & 87 & Pick up the left connector and insert it into the port. \\
38 & Grasp the knife and cut the cucumber at every scored plane (mesh C). & 88 & Pick up the gear and insert it into the painted green base. \\
39 & Grasp the knife and cut the cucumber at every scored plane (mesh D). & 89 & Pick up the gear and insert it into the steel base. \\
40 & Grasp the knife and cut the cucumber at every scored plane (mesh E). & 90 & Pick up the gear and insert it into the right base. \\
41 & Grasp the knife and cut the pickling cucumber at every scored plane. & 91 & Pick up the gear and insert it into the middle base. \\
42 & Grasp the knife and cut the eggplant at every scored plane. & 92 & Pick up the gear and insert it into the left base. \\
43 & Grasp the knife and cut the garlic bread at every scored plane. & 93 & Pick up the white painted gear and insert it into the base. \\
44 & Grasp the knife and cut the lotus root at every scored plane. & 94 & Pick up the painted green gear and insert it into the base. \\
45 & Grasp the knife and cut the okra at every scored plane. & 95 & Pick up the steel gear and insert it into the base. \\
46 & Grasp the knife and cut the parsnip at every scored plane. & 96 & Pick up the right gear and insert it into the base. \\
47 & Grasp the knife and cut the sweet potato at every scored plane. & 97 & Pick up the middle gear and insert it into the base. \\
48 & Grasp the knife and cut the zucchini at every scored plane (mesh A). & 98 & Pick up the left gear and insert it into the base. \\
49 & Grasp the knife and cut the zucchini at every scored plane (mesh B). & 99 & Pick up the medium gear and insert it onto the matching shaft of the gear base. \\
\bottomrule
\end{tabularx}
\end{table*}

\section{Tac-LeWM Pretraining}

\noindent\textbf{Heterogeneous Pretraining Data.}
We pretrain Tac-LeWM on a heterogeneous visuo-tactile corpus to expose it to contact dynamics beyond a single sensor and visual domain.
The corpus comprises approximately 3.0M frames and 42 hours of physical interaction from the in-domain training demonstrations of \textsc{TacMind-Bench}, ToucHD~\cite{feng2026anytouch}, Daimon-Infinity~\cite{daimon_infinity_2026}, RDP~\cite{rdp}, VLA-Touch~\cite{vlatouch}, and VisuoTactile-QINGLOONG~\cite{baihu_vtouch_2026}; all evaluation rollouts and OOD perturbation settings are excluded.
These sources cover three vision-based tactile sensors, GelSight Mini, DIGIT, and Daimon DM-Tac, as well as diverse camera configurations, robot morphologies, and teleoperation interfaces; together, they provide broader contact and visual variation than \textsc{TacMind-Bench} alone.
We unify the observation interface by resizing visual inputs to a shared spatial resolution and arranging tactile observations in a consistent temporal layout.
The action interface must also accommodate distinct control parameterizations, including end-effector poses, wrist motions, joint-space commands, and gripper states.
We retain these source-specific semantics through a dedicated MLP action projection for each dataset before the shared dynamics predictor, without requiring conversion into a single kinematic space.
Within each dataset, we additionally normalize actions using statistics computed over valid action samples so that differences in physical scale or control parameterization do not dominate the shared predictor.

\noindent\textbf{Action-Aware Pretraining Objective.}
Visual and tactile history can explain many short-horizon state transitions, so minimizing latent prediction error alone need not make the predictor responsive to commanded actions.
Heterogeneous pretraining adds a supervision constraint: some source samples lack valid action labels.
We address these two issues by separating representation regularization from action-conditioned dynamics learning and adding an action-reliance margin.
Samples without valid action annotations contribute to the representation regularizer $\mathcal{L}_{\mathrm{rep}}=\lambda\,\mathrm{SIGReg}(Z^\tau)$ defined in the main paper, but are masked out of the prediction and action-reliance terms; missing actions are never padded with zeros.
For samples with action labels, the margin encourages the predictor to distinguish the true sequence $A_t^{+}$ from a randomly permuted alternative $\tilde{A}_t^{+}$ drawn within the same dataset.
Let $Z_t^{+}$ denote the target future tactile latent chunk over horizon $C$, and let $\hat{Z}_t^{+}$ and $\tilde{Z}_t^{+}$ denote predictions conditioned on the true actions $A_t^{+}$ and permuted actions $\tilde{A}_t^{+}$, respectively.
The complete pretraining objective is
\begin{equation}
\begin{split}
\mathcal{L}
&=
\mathcal{L}_{\mathrm{rep}}
+
\lambda_{\mathrm{pred}} m^{a}
\left\|
\hat{Z}_t^{+}-Z_t^{+}
\right\|_2^2\\
&\quad+
\lambda_{\mathrm{act}} m^{a}
\max\bigl\{0,\,
\mathcal{S}_{\mathrm{cos}}\left(\tilde{Z}_t^{+},Z_t^{+}\right)
\\
&\qquad\qquad
-
\mathcal{S}_{\mathrm{cos}}\left(\hat{Z}_t^{+},Z_t^{+}\right)
+\delta
\bigr\},
\end{split}
\end{equation}
where $m^{a}\in\{0,1\}$ indicates whether valid action supervision is available, $\mathcal{S}_{\mathrm{cos}}$ is the mean cosine similarity across the prediction horizon, and $\delta$ is a margin hyperparameter.
The action-reliance term encourages the true-action prediction to exceed the permuted-action prediction in similarity to the target by the prescribed margin.
We permute actions strictly within each dataset source so that identifying the source alone cannot satisfy this constraint.
When computing gradients for the action-reliance term, we detach the target representation $Z_t^{+}$ and the permuted conditioning branch; gradients from this comparison therefore pass only through the true-action prediction branch.
We also randomly drop auxiliary visual features during pretraining to encourage the tactile pathway to retain predictive contact dynamics when visual conditioning is absent.

\noindent\textbf{Pretraining Protocol.}
We assign the largest sampling weight in the unified mixture to the \textsc{TacMind-Bench} training demonstrations to emphasize the downstream manipulation domain.
RDP, VLA-Touch, VisuoTactile-QINGLOONG, ToucHD, and Daimon-Infinity complement this domain coverage with additional contact dynamics and cross-sensor diversity.
Together, source-specific action projections and per-source normalization accommodate heterogeneous controls, while masked supervision and the action-reliance margin determine which samples train action-conditioned dynamics and how the predictor uses their actions.

\begin{table*}[t]
\centering
\caption{Ablation study of Tac-LeWM architectural components on linear probing.}
\label{tab:taclewm_ablation_probe}
\small
\setlength{\tabcolsep}{3pt}
\begin{tabular}{lcccc}
\toprule
Variant
& Proprio $\text{MSE}_z\downarrow$ / $r\uparrow$
& Gripper $\text{MSE}_z\downarrow$ / $r\uparrow$
& Action $\text{MSE}_z\downarrow$ / $r\uparrow$
& Pad $\text{MSE}_z\downarrow$ / $r\uparrow$ \\
\midrule
Full
& 0.1816 / 0.8981
& \textbf{0.1958} / \textbf{0.9053}
& 0.1841 / 0.8976
& \textbf{0.2771} / \textbf{0.7968} \\
w/o Camera
& 0.1748 / 0.9027
& 0.1958 / 0.9053
& 0.1770 / 0.9022
& 0.2905 / 0.7855 \\
w/o Causal Attn
& \textbf{0.1733} / \textbf{0.9035}
& 0.1958 / 0.9053
& \textbf{0.1756} / \textbf{0.9030}
& 0.3557 / 0.7282 \\
w/o CTR
& 0.1831 / 0.8971
& 0.1958 / 0.9053
& 0.1851 / 0.8968
& 0.3366 / 0.7476 \\
\bottomrule
\end{tabular}
\end{table*}

\section{Additional Ablation Study}

\noindent\textbf{Tac-LeWM Components.}
Causal temporal attention and CTR improve contact deformation representations and future prediction. The three interventions match the main-paper ablations: w/o Camera removes auxiliary visual conditioning from the predictor, w/o Causal Attn replaces temporal causal attention with single-frame self-attention, and w/o CTR replaces CTR with standard ViT raster positional encoding.
All variants perform comparably on instantaneous proprioception, gripper, and action probes (Table~\ref{tab:taclewm_ablation_probe}), consistent with single-frame contact geometry providing strong cues about current robot kinematics.
Removing causal attention or CTR nevertheless increases tactile pad deformation energy error, which indicates that temporal history and canonical spatial structure help capture fine-grained contact deformation.
The reconstruction results exhibit the same distinction (Table~\ref{tab:taclewm_ablation_recon}): current-frame quality remains similar, but future prediction loses nearly 6\,dB in PSNR without causal attention or CTR.
Together, these results establish the contribution of causal temporal aggregation and canonical directional encoding to forecasting future contact states.

\section{Inference Acceleration}
\label{sec:tacmind0-inference}

Compiling the action expert reduces the cost of the repeated ODE integration required by flow-reversal action refinement.
We compile the flow-matching expert with TensorRT and use CUDA Graphs to capture and replay static computation graphs.
This combination reduces GPU kernel launch overhead for both single-pass action generation and the three integration passes used by predictive tactile reflection.

We benchmark an 8-layer transformer action expert with hidden dimension 512 on an NVIDIA RTX~5090 GPU, reporting median latency over 30 runs (Table~\ref{tab:tacmind0-inference-latency}).
TensorRT with CUDA Graphs reduces single-pass integration latency from 115.4\,ms to 4.9\,ms and three-pass reflection latency from 300.9\,ms to 14.0\,ms.
Together with the 10.33\,ms per-step streaming tactile encoder, these timings correspond to approximately 40\,Hz reflective action generation when visual-language context encoding is amortized across action chunks.
The resulting latency supports the practical use of predictive tactile reflection in high-frequency closed-loop control.

\begin{table*}[t]
    \centering
    \small
    \caption{Action-expert integration latency in milliseconds on an NVIDIA RTX~5090 across 30 runs.}
    \label{tab:tacmind0-inference-latency}
    \begin{tabular}{lcc}
        \toprule
        Inference Backend & Single Pass in ms & Three-Pass Reflection in ms \\
        \midrule
        PyTorch                & 115.4 & 300.9 \\
        CUDA Graph             &  16.5 &  57.0 \\
        TensorRT               &  11.9 &  32.4 \\
        TensorRT + CUDA Graph  & \textbf{4.9} & \textbf{14.0} \\
        \bottomrule
    \end{tabular}
\end{table*}

\section{Real-World Experiment}
\label{sec:real_world}

\subsection{Experimental Setup}

\begin{figure*}[t]
    \centering
    \includegraphics[width=\textwidth]{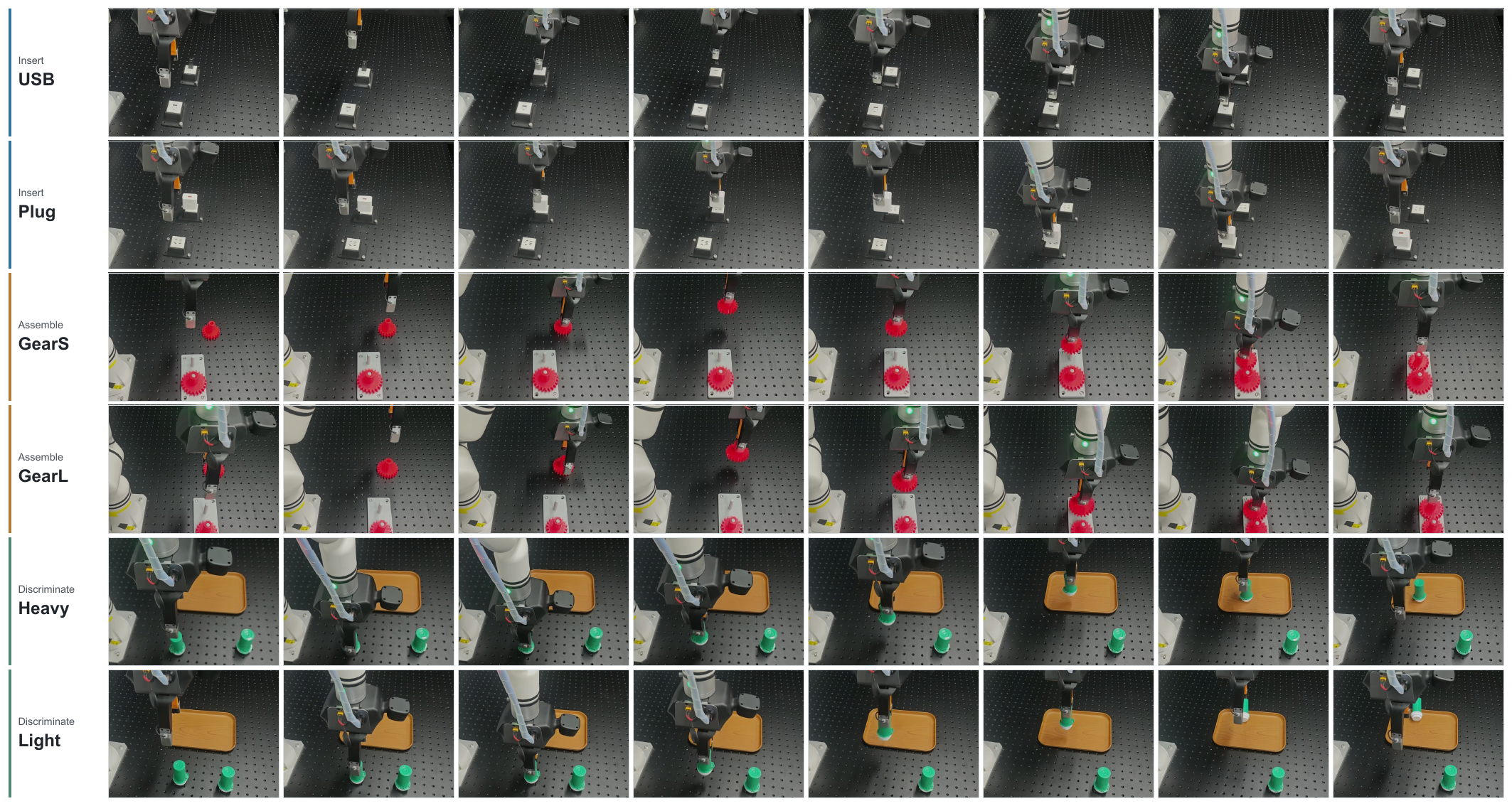}
    \caption{Representative rollouts for the six real-world manipulation subtasks. From top to bottom, the rows depict USB insertion, plug insertion, small-gear assembly, large-gear assembly, heavy-bottle discrimination, and light-bottle discrimination across eight chronological frames.}
    \label{fig:real_world_tasks}
\end{figure*}

We evaluate fine manipulation and tactile discrimination on a RealMan-ISF arm equipped with a Daimon DM-Tac W tactile sensor.
The physical benchmark contains six subtasks in three categories: \textit{Insert} tests tight-tolerance alignment, \textit{Assemble} tests contact-guided meshing, and \textit{Discriminate} tests physical weight discrimination.
\begin{itemize}
    \item \textbf{Insert USB} requires aligning a USB connector with its port and completing full insertion.
    \item \textbf{Insert Plug} requires aligning a power plug with its socket and completing insertion.
    \item \textbf{Assemble Small Gear (GearS)} requires aligning a small gear with a target shaft and completing assembly through positional and rotational adjustments.
    \item \textbf{Assemble Large Gear (GearL)} requires aligning a large gear with a target shaft and completing assembly through positional and rotational adjustments.
    \item \textbf{Discriminate Heavy Bottle} requires identifying and grasping a bottle containing an added internal weight through physical interaction.
    \item \textbf{Discriminate Light Bottle} requires identifying and grasping a lighter bottle without an added weight through physical interaction.
\end{itemize}
Figure~\ref{fig:real_world_tasks} shows representative rollouts for all six subtasks, with eight chronological frames per row from initial approach to task completion.
We collect 50 teleoperated demonstrations per subtask, for 300 real-world trajectories in total, and evaluate each method in 20 physical trials per subtask.

\subsection{Implementation Details}

We compare TacMind-0 with two generalist baselines: the vision-only VLA model OpenPI0.5~\cite{pi05} and the visuo-tactile model N0-VTLA~\cite{team2026n_0}.
All models are trained for 40,000 steps on 8 NVIDIA A100 GPUs with a global batch size of 64.
To account for differences in convergence speed, we evaluate checkpoints from 20,000, 30,000, and 40,000 steps using validation rollouts, then report the selected checkpoint's test success rate over the 20 physical trials per subtask.

\begin{figure*}[t]
    \centering
    \includegraphics[width=\textwidth]{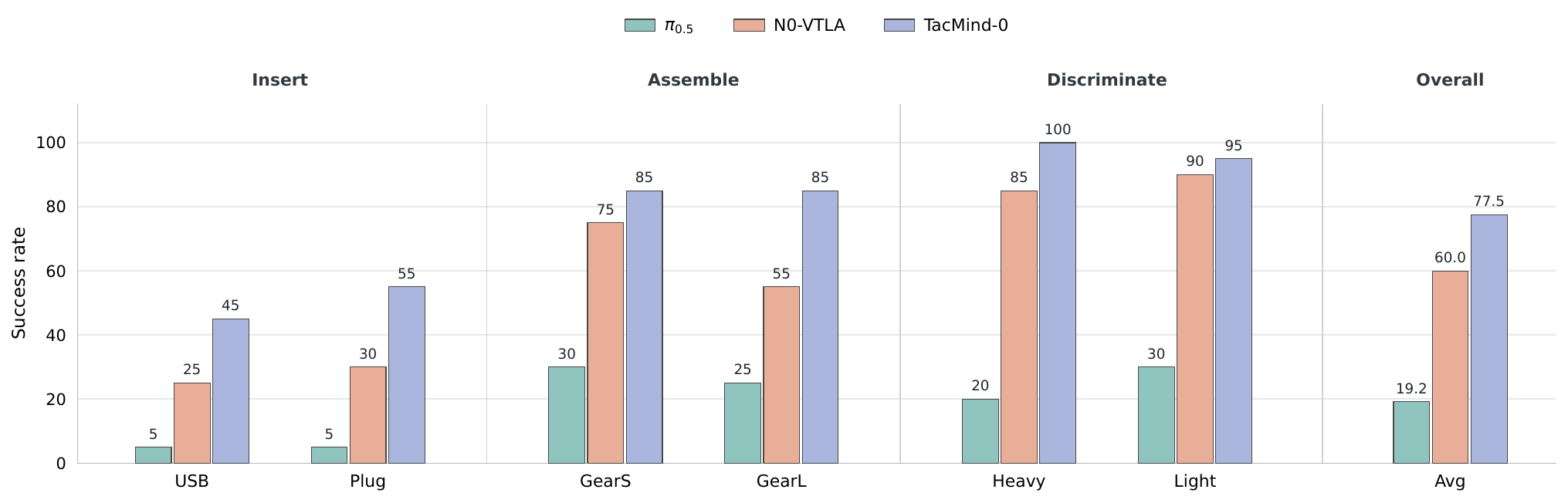}
    \caption{TacMind-0 improves success across all six real-world subtasks. We compare OpenPi0.5, N0-VTLA, and TacMind-0 over 20 physical trials per task.}
    \label{fig:real_world_success_rates}
\end{figure*}

\subsection{Task Analysis}

TacMind-0 achieves 77.5\% average success across the six physical subtasks, compared with 60.0\% for N0-VTLA and 19.2\% for OpenPi0.5 (Fig.~\ref{fig:real_world_success_rates}).
The improvements span insertion, assembly, and weight discrimination.

\noindent\textbf{Insert.}
TacMind-0 achieves the highest success rate on both USB and Plug insertion, while N0-VTLA improves over the rarely successful vision-only OpenPi0.5 baseline.
These tasks require millimeter-level alignment and immediate corrections when contact reveals a positional offset, including obstructions and corrective directions that are difficult to infer visually.
N0-VTLA supplies tactile tokens for this feedback, whereas TacMind-0 combines a streaming tactile state with predictive reflection to guide post-contact adjustments.
The performance gains support the effectiveness of this combined design for precise insertion.

\noindent\textbf{Assemble.}
TacMind-0 maintains high success across both gear sizes, while OpenPi0.5 struggles and N0-VTLA performs worse on the larger gear GearL.
Gear assembly requires locating a visually occluded shaft and regulating rotational meshing after contact.
The consistent performance across sizes supports TacMind-0's ability to use predictive tactile feedback for contact localization and rotational alignment.

\noindent\textbf{Discriminate.}
TacMind-0 achieves near-perfect success on both Heavy and Light bottle discrimination, which requires active probing to infer internal weight before the final lift.
The vision-only OpenPi0.5 baseline succeeds in fewer than half of trials on either task; completion requires both identifying the target bottle and lifting it stably.
N0-VTLA improves discrimination with tactile feedback, and TacMind-0's further gains demonstrate reliable physical state disambiguation on real robot hardware.

\end{document}